\def\showauthors{}
\newif\ifanon
\ifdefined\showauthors
  \anonfalse
  \AtBeginDocument{\iclrfinalcopy}
\else
  \anontrue
\fi
\documentclass{article}
\usepackage{iclr2027_conference,times}
\ClearShipoutPicture

\usepackage[utf8]{inputenc}
\usepackage[T1]{fontenc}
\usepackage{hyperref}
\usepackage{url}
\usepackage{amsmath,amssymb,amsthm}
\usepackage{booktabs}
\usepackage{graphicx}
\usepackage{xcolor}
\usepackage{algorithm}
\usepackage{algpseudocode}
\newcommand{\rev}[1]{\textcolor{black}{#1}}
\newcommand{\bluechange}[1]{\textcolor{black}{#1}}
\newenvironment{revblock}{\color{black}}{}
\usepackage{needspace}
\usepackage{wrapfig}
\usepackage[font=small,labelfont=rm,skip=4pt]{caption}

\ifanon
  \hypersetup{colorlinks=true,linkcolor=black,citecolor=black,urlcolor=black,
              pdfauthor={},pdftitle={},pdfcreator={},pdfproducer={}}
\else
  \hypersetup{colorlinks=true,linkcolor=black,citecolor=black,urlcolor=black,
              pdfauthor={Xi Lin, Feihong Zhang, Yulong Shi, Yanghong Mei, Zuxing Lu, Xiaofan Zhu, Zihao Liang, Zhirui Gao, Zhaowen Li},
              pdftitle={ReSync: Re-Aligning the Two Clocks of Asynchronous World-Action Models},
              pdfcreator={},pdfproducer={}}
\fi
\usepackage{makecell}
\usepackage{microtype}
\usepackage{placeins}
\usepackage{afterpage}

\usepackage{etoolbox}
\pretocmd{\section}{\needspace{3\baselineskip}}{}{}

\newcommand{\lead}[1]{\needspace{4\baselineskip}\paragraph{#1}}

\newtheorem{proposition}{Proposition}

\title{ReSync: Re-Aligning the Two Clocks of\\Asynchronous World-Action Models}

\author{
\href{https://xilin03.github.io/}{Xi Lin}$^{1,*,\dagger}$ \quad
Feihong Zhang$^{2,*}$ \quad
Yulong Shi$^{4,*}$ \quad
Yanghong Mei$^{3}$ \quad
Zuxing Lu \\
Xiaofan Zhu$^{4}$ \quad
Zihao Liang$^{4}$ \quad
Zhirui Gao$^{4}$ \quad
Zhaowen Li$^{4,\ddagger}$ \\
{\small $^{1}$Johns Hopkins University \quad $^{2}$Tsinghua University} \\
{\small $^{3}$University of Chinese Academy of Sciences \quad $^{4}$Yinwang Intelligent Technology Co., Ltd.}
}

\begin{document}
\maketitle
\makeatletter\global\@topnum=-1\makeatother
\raggedbottom
\ifanon\else
  \fancyhead[L]{Under review as a conference paper at ICLR 2027}
  \renewcommand{\headrulewidth}{0.4pt}
  \begingroup
  \renewcommand\thefootnote{}
  \footnotetext{\fontsize{7.5}{9}\selectfont
    $^{*}$Equal contribution.\quad
    $^{\dagger}$Project leader: \href{mailto:xilin03@outlook.com}{xilin03@outlook.com}.\quad
    $^{\ddagger}$Corresponding author: \href{mailto:lizhaowen3@huawei.com}{lizhaowen3@huawei.com}.}
  \addtocounter{footnote}{-1}
  \endgroup
\fi

\begin{abstract}
\looseness=-1
\rev{Joint world--action models often denoise actions in a few steps while refining future video for many more. The asymmetry keeps control fast and the predicted future sharp, but it also creates two inference clocks: an action can become executable before the world evidence that should justify it has resolved. We use the \emph{commitment--evidence gap} to diagnose this mismatch. Across a controlled family of X-WAM sampling schedules, widening the gap reverses the preferred use of test-time compute: candidate selection loses value while advancing world evidence gains it, and the reversal persists when candidate selection receives the same $3.40\times$ budget as our method. \textbf{ReSync} turns this observation into an inference rule. It pauses the action at a point the model was trained to support, advances only the world, and then resumes the native action schedule, adding no parameters and no action-denoising updates. On a frozen 960-rollout RoboCasa panel, ReSync improves X-WAM from 80.31\% to 84.79\%. At the same $3.40\times$ forward budget, five natural alternatives (waiting, resampling candidates, extending action denoising, synchronizing the two streams, and warm-starting the world) reach between 80.52\% and 82.92\%, and a fully synchronized schedule at $5.00\times$ reaches 83.23\%. Reaching the same world state in fewer, larger steps keeps most of the gain (83.85\%) at $1.80\times$ compute. Fixed on the native schedule, the rule also selects strong operating points on two held-out schedules, and it improves RoboTwin 2.0 by 2.67\,pp and a two-clock Cosmos Policy interface by 4.38\,pp. In our experiments, which clock the extra computation advances matters more than how much of it there is.}
\end{abstract}

\begin{figure}[!htbp]
  \centering
  \includegraphics[width=\textwidth]{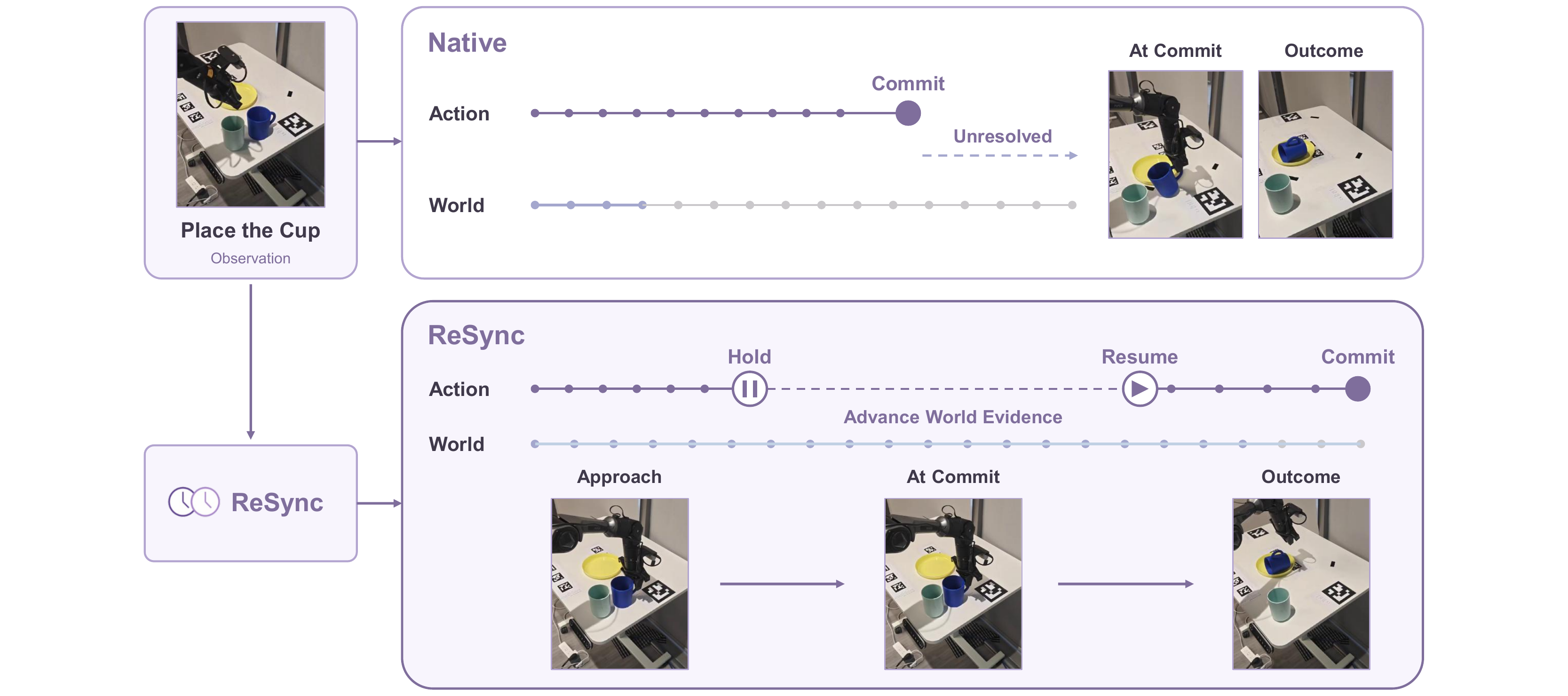}
  \caption{\textbf{Action commitment can precede useful world evidence.}
  Native X-WAM commits after the ten-step action schedule finishes; ReSync
  holds that action state, advances the world inside supported denoising time,
  and resumes native action denoising before commitment. Photographs are
  recorded execution frames; blur indicates evidence resolution schematically
  and is not model-decoded output.}
  \label{fig:teaser}
\end{figure}

\ifanon\clearpage\fi

\section{Introduction}

World-action models make an appealing promise for manipulation: the policy
predicts a future and acts inside that prediction with one generative model.
The strongest systems often give those two outputs different denoising
schedules. Actions are decoded quickly because control is latency-sensitive;
future video is refined longer because a world prediction is useful only if it
is sharp enough to carry task evidence. The price is easy to miss: action and
world no longer run on the same clock.

\looseness=-1
The robot reaches for a cup. Its action is already executable, while the
imagined future that should justify the reach is still taking shape. Nothing has
failed to generate. The decision has simply arrived before the evidence it should have waited for.

\looseness=-1
X-WAM makes this timing structure explicit: the action stream finishes in ten
updates while the world follows a 50-step trajectory \citep{guo2026xwam}.
Native execution therefore commits after only about one fifth of the world
schedule has elapsed. We call the mismatch the
\textbf{commitment--evidence gap}. The commitment clock measures progress toward
an executable action; the evidence clock measures progress toward a resolved
future.

\begin{quote}
\looseness=-1
\textbf{Two-clock principle.}
When commitment and evidence run on separate clocks, useful test-time compute makes evidence available while the action can still change.
\end{quote}

\looseness=-1
\rev{This principle separates three ideas that are easy to conflate. The \emph{gap} is a diagnostic: it says how far the evidence clock lags at commitment. The \emph{crossover} is an empirical finding: across a controlled family of schedules from one checkpoint, candidate selection is the better use of compute at small gaps and advancing world evidence at larger ones, with the order flipping between the measured $G=0.44$ and $G=0.58$ schedules even when selection receives the same $3.40\times$ budget. \emph{ReSync} is the method: once the timing problem is visible, it reads the native schedules, plus one fixed measurement of how many action updates must remain, to decide where world-only computation can be inserted.}

\rev{The matched-compute experiments show why this distinction matters. Running the action longer does not reproduce the gain. Neither does spending the same budget on more candidates, synchronizing the two streams, or sharpening the world before joint denoising; each remains significantly below ReSync. Even Sync-50, which denoises both streams for 50 steps, uses $5.00\times$ native compute and still trails ReSync at $3.40\times$. The world warm-start control is especially revealing: it predicts the future as accurately as ReSync, measured against what actually happens, yet ends lower. Useful evidence has to arrive at a joint state the model was trained to couple, and early enough that the remaining action updates can absorb it.}

\rev{That observation turns the question from ``how much more should we denoise?'' into ``when can extra world computation still change the decision?'' Two placement constraints come directly from the sampler and the training schedule: the pause must happen at an action state the sampler actually visits, and the world may not run further ahead of the action than training ever showed. A third, the number of action updates left after the pause, is fixed once by a coarse placement scan and then held constant. Together they determine the ReSync operating point.}

\looseness=-1
\rev{On the primary RoboCasa panel, this rule raises X-WAM from 80.31\% to 84.79\%. The gain is spread across 17 of 24 tasks and every perturbation group. A world-advance sweep rises to the supported $M=24$ point and then falls, and on two held-out schedules the peak moves with the support boundary, so the useful region is interior rather than ``more world compute is better.'' The cost is an explicit number of extra forwards, and reaching the same world state in eight larger steps keeps most of the gain at $1.80\times$ compute. The same rule transfers unchanged to RoboTwin and, after applying the same action-recovery convention to a constructed two-clock interface, to Cosmos Policy, where it also beats native synchronous inference while using fewer forwards. Across these results, the recurring pattern is simple: \textbf{world prediction is valuable when it becomes available while the action can still use it.}}

\section{Related Work}

\looseness=-1
\rev{Predictive policies couple control to future generation through video or joint world--action latents \citep{du2023unipi,wu2023gr1,cheang2024gr2,hu2024vpp,li2025uva,zhang2025dreamvla}. X-WAM and Cosmos Policy expose the two-clock coupling studied here \citep{guo2026xwam,kim2025cosmospolicy}. Long-WAM extends the observation context of such models and pairs it with inference optimization and asynchronous chunk execution for real-time control \citep{huang2026longwam}.} Utility-WAM corrects action changes induced by the predicted future \citep{lin2026utilitywam}. \rev{Candidate and search methods spend test-time compute on alternatives or verification \citep{kwok2025robomonkey,zhao2026adv,zhao2026geobon,chen2026caps}; timestep and cache methods redistribute sampling effort \citep{pan2024tstitch,sun2026sants,huang2026noisegate,wang2026evo,li2026elastic}. ReSync freezes the existing model, keeps the native number of action updates, and uses its joint timestep support to move world computation into the part of inference where the action can still respond. Appendix~\ref{app:alloc} compares the mechanisms directly.}

\section{The Two Clocks: Diagnosis and Empirical Crossover}
\label{sec:gap}

\subsection{The gap as a diagnostic}

\looseness=-1
A WAM that generates actions and future observations has two conceptually
different progress variables. The \emph{commitment clock} measures progress
toward an executable action; the \emph{evidence clock} measures progress toward
a resolved world prediction. A synchronous sampler ties the two together. An
asynchronous sampler does not.

\begin{figure}[!tbp]
  \centering
  \includegraphics[width=\textwidth]{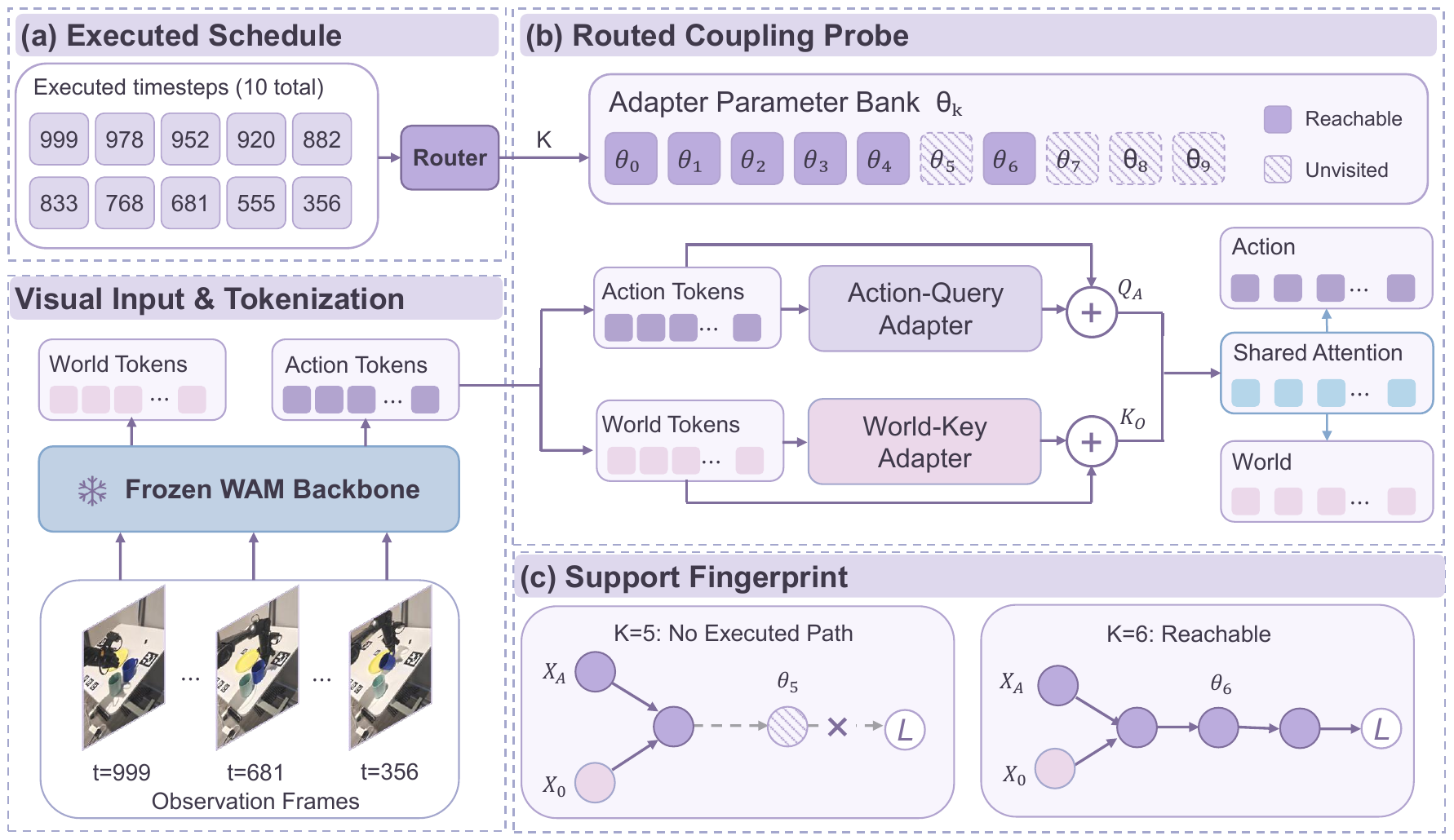}
  \caption{\rev{\textbf{Where world evidence enters the action.} (a) X-WAM's ten executed action timesteps fall into only six of ten timestep bins. (b) World and action tokens share one frozen DiT; at every joint update, action queries attend to the current, partially denoised world keys, so the action can use only what the world has resolved at that moment. To make this coupling visible, a lightweight probe adds bin-specific adapters to the action-query and world-key projections. (c) Only bins reached by the executed schedule learn: no executed timestep reaches bin 5, while $t=356$ reaches bin 6, and the pattern matches all ten bins across three seeds; ReSync trains nothing.}}
  \label{fig:arch}
\end{figure}

\rev{Figure~\ref{fig:arch} shows where the two clocks meet. World and action tokens are denoised by one shared DiT, and at every joint update the action tokens read the current, partially denoised world tokens through attention. Whatever the world has resolved at that moment is the evidence the action can use. The sampler also visits only some action timesteps, and only those carry this coupling (Fig.~\ref{fig:arch}a,c). Both facts shape ReSync: the pause must happen at an action state the sampler actually visits, and the world must be advanced before the action finishes.}

\begin{revblock}
Let $\tau_c$ denote action commitment. What matters is how much decision-relevant world evidence has resolved by $\tau_c$; the sampler gives a direct timing proxy, the fraction of world updates completed at commitment:
\[
\hat r_{\mathrm{sched}}(\tau_c)=\frac{n_O(\tau_c)}{T_O},
\qquad
G(\tau_c)=1-\hat r_{\mathrm{sched}}(\tau_c).
\]
We call $G$ the \textbf{commitment--evidence gap}. It is read off the schedule before any rollout and requires no task reward, critic, or additional model; Section~\ref{sec:operating_surface} checks it against directly measured prediction quality. X-WAM executes ten action updates against a 50-step world trajectory \citep{guo2026xwam}, so native commitment occurs after $10/50=0.20$ of the world updates, and $G=0.80$.
\end{revblock}

\rev{We measure whether that unresolved state can actually rank candidate actions with a within-set identifiability statistic $\kappa$.} For a
representation score $r$, let $\mathrm{AUC}_{\mathrm{within}}(r)$ be the AUC
against closed-loop candidate outcome computed only within candidate sets. For
the tested representation family $\mathcal{R}$,
\[
\kappa=
\frac{\max_{r\in\mathcal{R}}\mathrm{AUC}_{\mathrm{within}}(r)-\mu_0}
     {1-\mu_0},
\]
where $\mu_0$ is the mean of the familywise permutation-max null: each null
draw permutes outcomes within candidate sets, recomputes every score in
$\mathcal{R}$, and then takes the maximum AUC. Thus $\kappa=0$ places the best
tested representation at its familywise null level, while $\kappa=1$ denotes
perfect within-set separation. The screen fixes 80 candidate sets over 22 tasks
(2,560 trajectories), the representation family, and 10,000 permutations across
schedules; Appendix~\ref{app:prediction_details} gives the full protocol.

Holding candidates and scorer fixed, advancing only the world before
commitment raises $\kappa$ from 0.0187 to 0.063, 0.118, and 0.171.
\rev{Section~\ref{sec:pred1} tests the same ordering from the other direction, by changing the schedule instead of the intervention.}

\subsection{An empirical crossover in scaling value}
\label{sec:pred1}

\begin{figure}[!tbp]
  \centering
  \includegraphics[width=\textwidth]{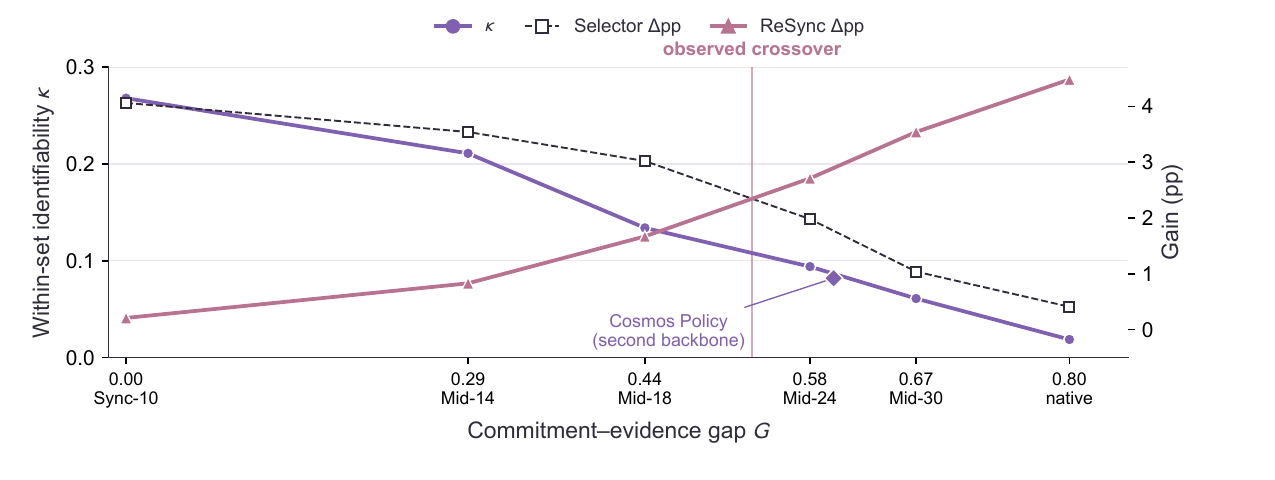}
  \caption{\textbf{The preferred compute axis reverses as the gap grows.}
  Larger gaps coincide with lower $\kappa$ and smaller selector gains, but
  larger evidence-advancement gains. The reversal lies between the measured
  $G=0.44$ and $G=0.58$ schedules; the marker interpolates.}
  \label{fig:gaplaw}
\end{figure}

\looseness=-1
\rev{We construct six inference schedules from one frozen X-WAM checkpoint, fixing the ten-step action schedule and the evaluation protocol while changing only the world-schedule length $T_O$ (the number in each setting name). At the largest gap, candidate utility is nearly unidentifiable ($\kappa=0.0187$). As the world resolves further before commitment, $\kappa$ and selector gains rise, while the gain from advancing evidence falls (Fig.~\ref{fig:gaplaw}).}

\rev{The selector uses one fixed label-free scorer in every row, distinct from the learned selector in Appendix~\ref{app:selector}. We report it at three budgets: the original 32-candidate bank, exactly the compute ReSync uses on that schedule, and a common $3.40\times$, the ReSync budget at the native schedule. The two smaller budgets use $N=4$ candidates on a fixed random fraction of decisions.}

\begin{table}[!htbp]
\centering
\captionsetup{font=footnotesize,skip=3pt}
\caption{\textbf{The crossover across six schedules of one checkpoint.} Only the world-schedule length changes. Selector gains use one fixed label-free scorer: with the original 32-candidate bank, with exactly the compute ReSync uses on that schedule, and with a common $3.40\times$. All gains are in pp over the schedule's own BASE; ReSync's budget is given in parentheses.}
\label{tab:gaplaw}
\footnotesize
\renewcommand{\arraystretch}{0.88}
\setlength{\aboverulesep}{0.8pt}
\setlength{\belowrulesep}{0.8pt}
\begin{tabular*}{\textwidth}{@{\extracolsep{\fill}}lrrrrrrr@{}}
\toprule
 & & & & \multicolumn{3}{c}{Selector $\Delta$pp} & \\
\cmidrule(lr){5-7}
Setting & $G$ & BASE & $\kappa$ & \thead{32-cand.\\bank} & {\color{black} \thead{same\\budget}} & {\color{black} \thead{$3.40\times$}} & {\color{black} \thead{ReSync $\Delta$pp\\(budget)}} \\
\midrule
Sync-10        & 0.00 & 76.15 & 0.268  & +4.06 & {\color{black} 0.00} & {\color{black} +2.19} & {\color{black} +0.21$^\dagger$ (1.00$\times$)} \\
Mid-14         & 0.29 & 79.48 & 0.211  & +3.54 & {\color{black} +0.21} & {\color{black} +2.08} & {\color{black} +0.83 (1.20$\times$)} \\
Mid-18         & 0.44 & 81.04 & 0.134  & +3.02 & {\color{black} +0.42} & {\color{black} +1.77} & {\color{black} +1.67 (1.40$\times$)} \\
Mid-24         & 0.58 & 81.25 & 0.094  & +1.98 & {\color{black} +0.52} & {\color{black} +0.94} & {\color{black} +2.71 (1.80$\times$)} \\
Mid-30         & 0.67 & 80.73 & 0.061  & +1.04 & {\color{black} +0.31} & {\color{black} +0.52} & {\color{black} +3.54 (2.20$\times$)} \\
X-WAM (native) & 0.80 & 80.31 & 0.0187 & +0.41 & {\color{black} +0.21} & {\color{black} +0.21} & {\color{black} +4.48 (3.40$\times$)} \\
\bottomrule
\end{tabular*}

{\footnotesize $^\dagger$ At Sync-10 ReSync inserts no world-only step and equals BASE; the +0.21\,pp is run-to-run variation.}
\end{table}

\bluechange{Under equal compute on each schedule, advancing the world is the better use of that compute wherever a gap exists. With the selector given $3.40\times$ on every schedule, at least as much as ReSync ever uses, selection still wins at small gaps and the order flips between Mid-18 and Mid-24, the same place as with the 32-candidate bank. The flip is therefore stable once selection has a few candidates' worth of compute: selection keeps scaling with more candidates, whereas world advancement is capped by the joint timestep regime the model was trained on. We treat it as an empirical regime marker rather than an exact threshold.}

\section{Method: ReSync}
\label{sec:method}

The gap tells us whether evidence is late; the crossover tells us which compute
axis paid off in the controlled schedule family. Neither specifies an
intervention. ReSync is the step that turns those observations into an
inference rule. It asks where world-only computation can be inserted so that
the resulting world/action state is supported and the action still has time to
change.

\begin{figure}[!tbp]
  \centering
  \includegraphics[width=\textwidth]{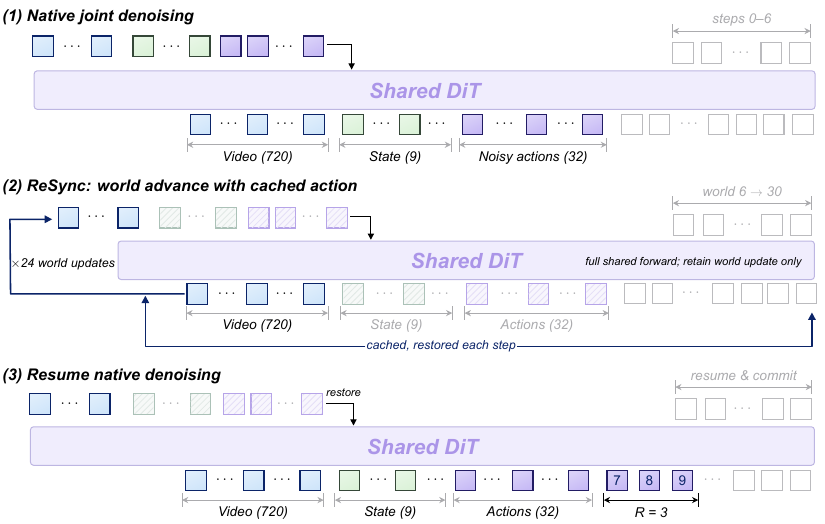}
  \caption{\rev{\textbf{ReSync advances only the evidence clock inside the usable window.} The action state is held, the world advances, and native action denoising resumes while enough action updates remain.}}
  \label{fig:method}
\end{figure}

\subsection{Where world computation can be inserted}

\looseness=-1
\rev{The gap is the gate: if evidence is already resolved at commitment, there is little reason to move world computation earlier. Once a gap is present, three constraints determine where it can work.}

\rev{\textbf{Pause on a visited action state.} The pause is placed only at action states that native inference actually visits, $j\in\mathcal{S}_{\mathrm{exec}}$ (Fig.~\ref{fig:arch}a).}

\rev{\textbf{Leave the action time to respond.} Extra evidence matters only if native action denoising can still absorb it. A coarse placement scan fixes the number of action updates that must remain after the pause, $R_{\min}=3$, once; the same value is then used on every schedule, benchmark, and backbone.}

\begin{revblock}
\textbf{Stay inside the training regime.} X-WAM's asynchronous noise sampling constructs the noisy world time as the action time plus a nonnegative residual (Sec.~3.3 of \citet{guo2026xwam}), so in training the world is never less noisy than the action: $t_O\ge t_a$. With $\psi(j)$ the world-schedule index reached after $j$ joint updates, we therefore advance the world only while this still holds:
\[
M_{\max}(j)=\max\{m:t_O^{\psi(j)+m}\ge t_a^j\}.
\]
\end{revblock}

\begin{revblock}
Together these constraints define the feasible placements
\[
\mathcal{F}=\{(j,M):j\in\mathcal{S}_{\mathrm{exec}},\;J-1-j\ge R_{\min},\;0\le M\le M_{\max}(j)\}.
\]
The schedule supplies the first and third constraints; the placement scan supplies the single value of $R_{\min}$. None of these quantities is retuned in any later evaluation.
\end{revblock}

\noindent\textbf{Why not just shorten the world schedule?}
Table~\ref{tab:gaplaw} shows why. BASE success is non-monotonic in $T_O$:
shortening the world schedule reduces the gap, but beyond a point it also
removes useful generative refinement. ReSync keeps the longer world trajectory
and changes when a supported portion of it becomes available to the action.

\subsection{The ReSync operator}

The operator itself is almost disappointingly simple, and that is the point: the schedule has already made the hard allocation decision. \rev{ReSync reads the native world and action schedules, computes $M_{\max}(j)$, and takes the latest placement that still leaves $R_{\min}=3$ action updates, advancing the world as far as the training regime allows. At that placement it caches the action latent and its timestep, runs ordinary shared-DiT forwards while keeping only the world update, and restores the cached action after each forward. After $M$ such updates, native action denoising resumes from the saved state and commits after its remaining updates. The action therefore sees a more resolved world without an extra action-denoising step, parameter, critic, selector, or verifier; Algorithm~\ref{alg:resync} in Appendix~\ref{app:impl} gives the procedure.}

\subsection{Operating point}

\begin{revblock}
For X-WAM, $J=10$ and
\[
M_{\max}(j)=[0,4,8,12,16,20,24,28,32],\qquad j=0,\ldots,8.
\]
Requiring three remaining action updates removes $j\ge7$, leaving $(j^\star,M^\star)=(6,24)$ as the latest feasible point with the largest supported advance. Native X-WAM commits after 10 of 50 world updates; with 24 inserted world-only updates and the same ten action updates, the world has completed 34 of 50 updates at commitment.
\end{revblock}

The same action-recovery convention also defines our Cosmos interface. We
construct a 14-update action stream against its 35-step future-state schedule;
with $R_{\min}=3$, the latest safe cache is $j_C=10$. Cosmos uses a different
training construction from X-WAM, so the transfer interface fixes a ten-step
world-only advance, $M_C=10$, before closed-loop evaluation. All Cosmos results
below refer to this constructed $(J_C,j_C,M_C)=(14,10,10)$ interface. BASE and
ReSync execute the same 14 action updates.

\subsection{Compute and a lighter variant}

\looseness=-1
\rev{Each inserted X-WAM world-only update is one shared-DiT forward, so the default point costs $C=1+M/10=3.40\times$ native forward compute. Section~\ref{sec:controls} spends the same 34-forward budget in five other ways, making \emph{where} the compute goes the controlled variable. \bluechange{The rule fixes a target, how far along the world schedule the world should be when native action denoising resumes, rather than a step count. The same target can be reached in fewer, larger steps along the native world schedule: taking every sixth or every third of the 24 inserted timesteps gives $K=4$ or $K=8$ world-only forwards ($1.40\times$ or $1.80\times$). Native steps remain the default; Section~\ref{sec:cost} evaluates the lighter variants.}}

\section{Experiments}
\label{sec:exp}

\noindent\textbf{Benchmarks and protocol.}
The primary evaluation is a frozen clustered paired RoboCasa panel
\citep{nasiriany2024robocasa}: 192 physical-state clusters with five
policy-noise replicates each, for 960 rollouts across 24 tasks. Every method
receives the same physical states and policy-noise contract. Transfer uses a
300-rollout paired RoboTwin 2.0 panel \citep{chen2025robotwin2} and a
320-rollout RoboCasa panel with Cosmos Policy \citep{kim2025cosmospolicy}.
Clustered bootstrap intervals are the primary inferential quantity; exact paired
McNemar summaries are in Appendix~\ref{app:stats}.

\subsection{Does ReSync improve the policy?}

On the primary panel, native X-WAM succeeds on 80.31\% of rollouts. ReSync
raises success to 84.79\%, a +4.48\,pp change with 82 rescues and 39 breaks
(95\% cluster-bootstrap CI [+2.17,+6.74]). The gain is not carried by one task
family: 17 of 24 RoboCasa tasks improve, four are effectively unchanged, and
three decrease. Positive deltas appear under language, lighting, robot
placement, camera, and background perturbations. ReSync also composes with
candidate allocation, reaching 85.94\% at $4.15\times$ compute.

The main question is therefore not whether the intervention can help, but
whether its advantage is specific to \emph{where} the extra forwards are spent.

\subsection{Would the same compute help anywhere else?}
\label{sec:controls}

\looseness=-1
We spend the same 34 shared-DiT forwards in five obvious ways.
\textbf{wait-only} executes the extra forwards but restores both streams;
\textbf{matched resampling} buys additional candidate trajectories;
\textbf{Action-34} spends the budget on action denoising;
\textbf{Sync-34} uses a common 34-step schedule for both streams; and
\textbf{world warm-start} advances the world for 24 forwards before the native
ten joint updates. We also include \textbf{Sync-50}, the literal fully synchronized schedule, which is intentionally over budget at $5.00\times$.

\begin{table}[!htbp]
\centering
\captionsetup{font=small,skip=3pt}
\caption{\textbf{Where the compute goes matters.} The middle block spends the same 34 shared-DiT forwards ($3.40\times$) in six ways; the bottom block lists references at other budgets. CIs are paired cluster-bootstrap intervals against BASE. \bluechange{Holm $p$ tests each equal-budget alternative against ReSync, corrected over the five contrasts.} Rescue/break counts are in Appendix~\ref{app:stats}.}
\label{tab:main}
\small
\renewcommand{\arraystretch}{0.95}
\begin{tabular*}{\textwidth}{@{\extracolsep{\fill}}lrrrrr@{}}
\toprule
Method & Compute & SR (\%) & $\Delta$pp & 95\% CI & {\color{black}Holm $p$} \\
\midrule
BASE                & 1.00$\times$ & 80.31 & --    & --              & {\color{black}--} \\
\midrule
wait-only           & 3.40$\times$ & 80.52 & +0.21 & [$-$1.84,+2.19] & {\color{black}0.002} \\
matched resampling  & 3.40$\times$ & 82.08 & +1.77 & [$-$0.27,+3.83] & {\color{black}0.013} \\
Action-34           & 3.40$\times$ & 81.56 & +1.25 & [$-$0.84,+3.35] & {\color{black}0.005} \\
Sync-34             & 3.40$\times$ & 82.29 & +1.98 & [$-$0.10,+4.05] & {\color{black}0.019} \\
world warm-start    & 3.40$\times$ & 82.92 & +2.60 & [+0.31,+4.88]   & {\color{black}0.041} \\
\textbf{ReSync}     & 3.40$\times$ & \textbf{84.79} & \textbf{+4.48} & \textbf{[+2.17,+6.74]} & {\color{black}--} \\
\midrule
allocation $r25,N4$ & 1.75$\times$ & 82.71 & +2.40 & [+0.18,+4.61]   & {\color{black}--} \\
Sync-50             & 5.00$\times$ & 83.23 & +2.92 & [+0.71,+5.13]   & {\color{black}--} \\
ReSync + allocation & 4.15$\times$ & 85.94 & +5.63 & [+3.17,+8.12]   & {\color{black}--} \\
\bottomrule
\end{tabular*}
\end{table}

\begin{figure}[!tbp]
  \centering
  \includegraphics[width=\textwidth]{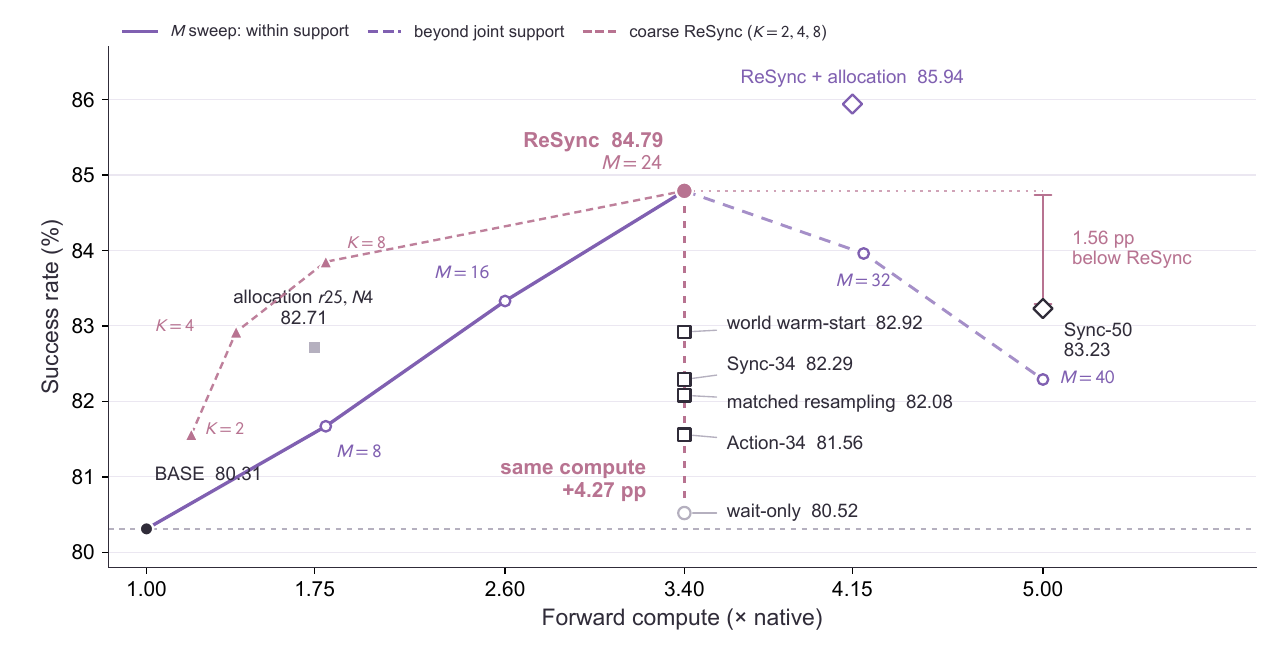}
  \caption{\rev{\textbf{One compute--success plane for every allocation.} World advancement (ReSync, $M=0$ to $40$) improves until the supported boundary and then gives the gain back; candidate allocation saturates earlier; the five equal-budget alternatives at $3.40\times$ and Sync-50 at $5.00\times$ all sit below ReSync's peak.}}
  \label{fig:pareto}
\end{figure}

\bluechange{No other use of the same 34 forwards comes close. The equal-budget alternatives top out at 82.92\%, 1.87\,pp below ReSync, and all five remain significantly lower after Holm correction across the five contrasts (adjusted $p\le0.041$; Table~\ref{tab:main}). The closest, world warm-start, also trails ReSync on both held-out schedules (Table~\ref{tab:heldout_schedules}); pooled over the three schedules, ReSync leads it by +1.63\,pp (95\% CI [+0.57,+2.73]).{\color{black}\footnote{Matched resampling allocates extra sampling on the closed-loop panel, whereas Table~\ref{tab:gaplaw} evaluates a fixed scorer on a frozen candidate bank; the two selection numbers therefore come from different candidate protocols.}}} \bluechange{Figure~\ref{fig:pareto} places all of these allocations on one compute--success plane.}

\rev{The controls separate several superficially similar ideas. Action-34 shows that extra action refinement is not enough. Warm-start shows that a sharper world \emph{before} joint denoising is not enough either: the world is moved ahead while the action is still near its initial noisy state, a timestep pairing the model never saw in training. ReSync waits until the action has progressed, advances the world inside the supported joint regime, and still leaves three native action updates to absorb that evidence.}

\rev{Sync-50 is the most direct way to remove the gap: let the action denoise for as long as the world does. It costs $5.00\times$ native compute and finishes 1.56\,pp below ReSync at $3.40\times$ (95\% CI [$-$0.28,+3.42]), so ReSync matches or exceeds the fully synchronized schedule with about a third less compute. Removing the gap is not the same as spending computation where the gap says it is useful.}

\subsection{Why is there a useful range at all?}
\label{sec:operating_surface}

\rev{If ReSync worked because more world denoising is always better, success would keep rising with $M$. It does not (Table~\ref{tab:msweep_main}). \bluechange{To see what the inserted steps actually change, we also measure the evidence directly on 384 paired decision states: at the moment the action is committed, we decode the world prediction and compare it with the frames that occur after executing that action, over the full frame (LPIPS) and over the gripper and target-object region given by simulator masks.}}

\begin{table}[!htbp]
\centering
\color{black}
\captionsetup{font=footnotesize,skip=3pt}
\caption{\rev{\textbf{Evidence improves up to the supported boundary; success follows only inside it.} World advance $M$ at $j^\star=6$; $M=24$ is the largest supported advance, and $M=32,40$ deliberately cross it. Warm-start spends 24 world-only forwards before joint denoising. Prediction quality compares the world prediction decoded at commitment with the frames observed after executing the committed action (384 paired decision states); task-region error is restricted to the gripper and target object and normalized by the native value; action change is the RMS difference from the native action in dataset standard-deviation units.}}
\label{tab:msweep_main}
\footnotesize
\renewcommand{\arraystretch}{0.88}
\begin{tabular*}{\textwidth}{@{\extracolsep{\fill}}lrrrrrrr@{}}
\toprule
 & \multicolumn{6}{c}{ReSync, world advance $M$} & \\
\cmidrule(lr){2-7}
 & 0 & 8 & 16 & \textbf{24} & 32 & 40 & {\color{black} warm-start} \\
\midrule
SR (\%)                      & 80.31 & 81.67 & 83.33 & \textbf{84.79} & 83.96 & 82.29 & {\color{black} 82.92} \\
$\Delta$pp                   & 0.00  & +1.35 & +3.02 & \textbf{+4.48} & +3.65 & +1.98 & {\color{black} +2.60} \\
{\color{black} LPIPS $\downarrow$}           & {\color{black} 0.422} & {\color{black} 0.379} & {\color{black} 0.343} & {\color{black} 0.309} & {\color{black} 0.303} & {\color{black} 0.317} & {\color{black} 0.312} \\
{\color{black} task-region error $\downarrow$} & {\color{black} 1.000} & {\color{black} 0.861} & {\color{black} 0.729} & {\color{black} 0.619} & {\color{black} 0.604} & {\color{black} 0.633} & {\color{black} 0.625} \\
{\color{black} action change}                & {\color{black} 0.000} & {\color{black} 0.084} & {\color{black} 0.128} & {\color{black} 0.172} & {\color{black} 0.237} & {\color{black} 0.291} & {\color{black} 0.209} \\
\bottomrule
\end{tabular*}
\end{table}

\bluechange{Up to the boundary, evidence and success rise together: the task-region error falls by 38\% from $M=0$ to $M=24$ while success rises by 4.48\,pp. Beyond it they separate. At $M=32$ the world is still slightly better than at $M=24$, yet success is lower ($M=24$ leads by +0.83\,pp, 95\% CI [+0.11,+1.57]); by $M=40$ the world itself starts to degrade. The decline therefore does not come from a worse world; it comes from pairing a world that has run past the training regime with an action that has not. World warm-start makes the same point from the other side: it reaches the same evidence as ReSync but delivers it before the action has formed, and ends 1.87\,pp lower. The boundary also moves with the schedule as the rule predicts: the best tested point is $M=16$ on 8/40, $M=24$ on 10/50, and $M=32$ on 12/60, with three action updates left after the pause in all three (Table~\ref{tab:heldout_schedules}).}

\begin{figure}[!tbp]
\centering
\begin{minipage}[b]{0.668\textwidth}\centering
\includegraphics[width=\linewidth]{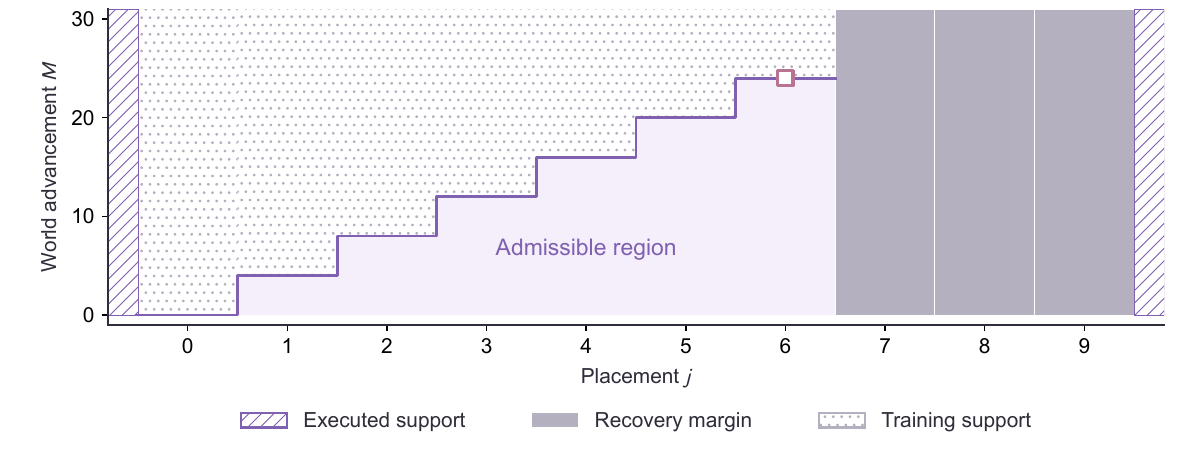}
\end{minipage}\hfill
\begin{minipage}[b]{0.312\textwidth}\centering
\includegraphics[width=\linewidth]{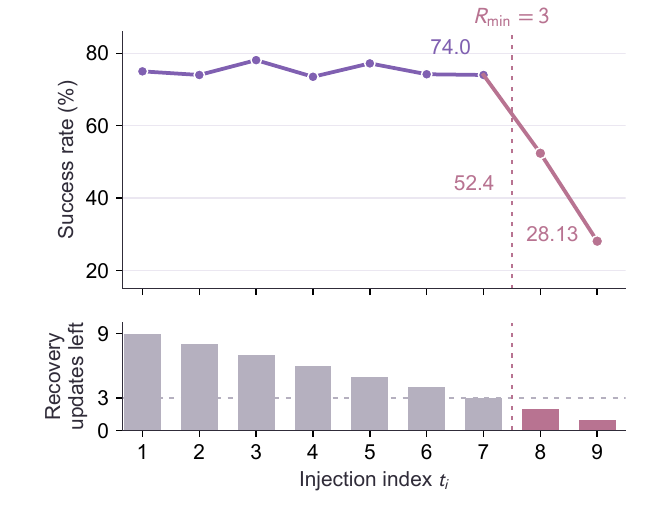}
\end{minipage}
\caption{\rev{\textbf{Two constraints carve out the high-return region.} Left: success over placement and world advance; shading marks support constraints, not success, and the schedule-selected $(6,24)$ point lies on the same plateau as the best safe cells. Right: success collapses when only one or two action updates remain after the pause.}}
\label{fig:window}
\end{figure}

\rev{Placement creates the other side of the window. Success remains stable through $t_i=7$ and drops sharply at $t_i=8,9$, where only two or one native action updates remain (Fig.~\ref{fig:window}, right). This coarse scan fixes $R_{\min}=3$ once. The full placement--advancement surface shows a broad high-return region inside the constraints: the schedule-selected $(6,24)$ point reaches 84.79\%, and the best observed safe cell reaches 85.10\%, on the same plateau, without being selected from that surface (Fig.~\ref{fig:window}, left).}

\color{black}
\subsection{What does the gain cost?}
\label{sec:cost}

\looseness=-1
\bluechange{The cost of ReSync is set by one number, the forwards spent on world-only updates (Table~\ref{tab:cost}). Reaching the default world target in fewer, larger steps keeps most of the gain: +2.60\,pp with four steps and +3.54\,pp with eight. With the same 18 forwards, eight native steps give +1.35\,pp, and candidate allocation gives +2.40\,pp at $1.75\times$. Adding allocation to the eight-step variant reaches 85.10\% at $2.55\times$, matching the default point with a quarter less compute.}

\begin{table}[!th]
\centering
\color{black}
\captionsetup{font=footnotesize,skip=3pt}
\caption{\bluechange{\textbf{Cost per policy decision.} Research code: one H800, as in all other experiments. Optimized: the same implementation for every method on four H800s at unchanged precision, P50 / P90 over 384 decisions after warm-up (Appendix~\ref{app:impl}). Each decision returns a 32-action chunk that runs for 1.6\,s. The $K$ variants reach the same world state as $M=24$ in fewer, larger steps. $^\dagger$Mean; a quarter of allocation's decisions run four candidates.}}
\label{tab:cost}
\footnotesize
\renewcommand{\arraystretch}{0.88}
\begin{tabular*}{\textwidth}{@{\extracolsep{\fill}}lrrrrrr@{}}
\toprule
Method & Forwards & Compute & Research code & Optimized (P50 / P90) & SR & $\Delta$pp \\
\midrule
X-WAM (native)              & 10   & 1.00$\times$ & 2.68\,s & 0.78 / 0.93\,s & 80.31 & -- \\
ReSync, $K=2$               & 12   & 1.20$\times$ & 3.20\,s & 0.93 / 1.10\,s & 81.56 & +1.25 \\
ReSync, $K=4$               & 14   & 1.40$\times$ & 3.70\,s & 1.08 / 1.27\,s & 82.92 & +2.60 \\
ReSync, $K=8$               & 18   & 1.80$\times$ & 4.73\,s & 1.38 / 1.57\,s & 83.85 & +3.54 \\
ReSync, $M=8$ native steps  & 18   & 1.80$\times$ & 4.74\,s & 1.40 / 1.61\,s & 81.67 & +1.35 \\
allocation $r25,N4$         & 17.5 & 1.75$\times$ & 4.81\,s$^\dagger$ & 0.82 / 3.38\,s & 82.71 & +2.40 \\
ReSync, $K=8$ + allocation  & --   & 2.55$\times$ & 6.82\,s & 1.40 / 3.83\,s & 85.10 & +4.79 \\
ReSync, $M=24$ (default)    & 34   & 3.40$\times$ & 8.97\,s & 2.61 / 3.02\,s & 84.79 & +4.48 \\
\bottomrule
\end{tabular*}
\end{table}

\bluechange{These are times per policy decision; each decision returns a 32-action chunk that runs for 1.6\,s at RoboCasa's 20\,Hz control rate. With asynchronous chunk execution \citep{huang2026longwam}, the next decision is computed while the current chunk runs, so a decision that finishes within 1.6\,s adds no waiting time. With research code, native X-WAM already exceeds this window, every method pauses, and the time ReSync adds remains visible. We therefore also ran every method with the same optimized implementation on four H800s at unchanged precision, whose actions match the research code to within $0.0082$ in normalized action space (Appendix~\ref{app:impl}). With it, native X-WAM and the $K$ variants up to $K=8$ finish within the window at the 90th percentile. In a replay of the full 960-rollout panel with these measured times, ReSync $K=8$ keeps its +3.54\,pp lead over native X-WAM and lengthens the median episode from 19.9\,s to 20.5\,s, while the default $M=24$ still pauses about 1.0\,s per decision (Table~\ref{tab:async}). For the lighter variants, the remaining cost is GPU compute rather than waiting time.}

\color{black}
\subsection{Does the rule work on schedules it has not seen?}
\label{sec:heldout_schedules}

\looseness=-1
The rule was fixed on the native 10/50 schedule. We now apply it unchanged to
two X-WAM schedule pairs that played no part in deriving it, using the same
frozen 960-rollout RoboCasa panel. With $R_{\min}=3$, the 8/40 pair gives the
latest safe cache $j^\star=4$ and the native timestep inequality gives
$M^\star=16$; the 12/60 pair gives $j^\star=8$ and $M^\star=32$. We record
those operating points before evaluating either local sweep.

\begin{table}[!htbp]
\centering
\captionsetup{font=footnotesize,skip=3pt}
\caption{\textbf{Frozen-rule validation on two unseen schedule pairs.}
For each pair we evaluate the rule-selected point, its $M\pm8$ neighbors, and
an equal-budget world warm-start; the $M+8$ point deliberately crosses the
schedule-derived support boundary. Cluster-bootstrap CIs compare the rule point with that schedule's BASE.}
\label{tab:heldout_schedules}
\footnotesize
\renewcommand{\arraystretch}{0.88}
\setlength{\aboverulesep}{0.8pt}
\setlength{\belowrulesep}{0.8pt}
\begin{tabular*}{\textwidth}{@{\extracolsep{\fill}}lrrrrrrrr@{}}
\toprule
\thead{Schedule\\$(J,T_O)$} & \thead{Rule\\$(j^\star,M^\star)$} &
BASE & \thead{$M^\star\!-\!8$\\SR} & \thead{Rule\\SR} &
\thead{$M^\star\!+\!8$\\SR} & \thead{Warm-start\\SR} &
\thead{Rule\\$\Delta$pp} & 95\% CI \\
\midrule
$(8,40)$  & $(4,16)$ & 79.58 & 82.29 & \textbf{83.65} & 83.02 & 82.19 & \textbf{+4.06} & [+1.72,+6.35] \\
$(12,60)$ & $(8,32)$ & 80.94 & 84.06 & \textbf{85.21} & 84.58 & 83.65 & \textbf{+4.27} & [+1.96,+6.61] \\
\bottomrule
\end{tabular*}
\end{table}

\looseness=-1
In both schedules the frozen rule lands at the highest of the three tested
ReSync advancement points, and the equal-budget warm-start remains lower. The
8/40 rule point uses $1+16/8=3.00\times$ native forward compute; the 12/60
point uses $1+32/12=3.67\times$. The same rule that explains the native 10/50
schedule therefore selects a high-return operating point before either new
local sweep is revealed.

\subsection{Does a fixed rule transfer?}

RoboTwin uses the same X-WAM asynchronous schedule, so the operating point
remains $(6,24)$. Cosmos changes both backbone and training construction. We
therefore transfer the action-recovery convention rather than X-WAM's
training-support inequality: the constructed 14/35 interface fixes
$(j_C,M_C)=(10,10)$ before evaluation. $R_{\min}$ and the number of action updates remain unchanged.

\begin{table}[!htbp]
\centering
\captionsetup{font=footnotesize,skip=3pt}
\caption{\textbf{The same timing rule transfers across benchmark and
backbone.} RoboTwin reuses the native X-WAM schedule; Cosmos uses the constructed 14/35 interface (14 action updates in both arms).}
\label{tab:transfer}
\footnotesize
\renewcommand{\arraystretch}{0.88}
\setlength{\aboverulesep}{0.8pt}
\setlength{\belowrulesep}{0.8pt}
\begin{tabular*}{\textwidth}{@{\extracolsep{\fill}}llrrrrrr@{}}
\toprule
Backbone & Benchmark & $G$ & $\kappa$ & $(j^\star,M)$ & BASE & ReSync & $\Delta$pp \\
\midrule
X-WAM & RoboCasa      & 0.80 & 0.0187 & $(6,24)$   & 80.31 & 84.79 & +4.48 \\
X-WAM & RoboTwin 2.0  & 0.80 & $-0.079^\ddagger$ & $(6,24)$ & 90.33 & 93.00 & +2.67 \\
Cosmos Policy & RoboCasa & 0.60 & 0.082 & $(10,10)$ & 66.25 & 70.63 & +4.38 \\
\bottomrule
\end{tabular*}

{\footnotesize $^\ddagger$ Matched 60-cluster subset; its uncertainty interval
spans zero.}
\end{table}

On RoboTwin, success rises from 90.33\% to 93.00\% (+2.67\,pp; 95\% CI
[+0.24,+5.06]). The 60-cluster estimate gives $\kappa=-0.079$ with an interval spanning zero,
so both RoboTwin and native RoboCasa fall in the practically non-identifiable
regime. Across benchmarks we use $G$ to identify that regime, while $\kappa$
remains a panel-specific estimate. RoboTwin also starts from a much higher
baseline and therefore has less absolute headroom.

\rev{On Cosmos, the constructed 14/35 interface gives $G=0.60$ and $\kappa=0.082$. Its released sampler supplies the 35-step future-state trajectory; on our 14-update action interface, the same three-update recovery convention places the pause at $j_C=10$, and we fix $M_C=10$ world-only forwards before evaluation. ReSync raises success from 66.25\% to 70.63\% (+4.38\,pp; 95\% CI [+1.06,+7.72]) while keeping the same 14 action updates, at $1+10/14=1.71\times$ the interface's forward compute. \bluechange{On the same panel, native synchronous Cosmos inference with 35 joint updates reaches 66.88\%. The constructed interface therefore matches native success within noise, and with ReSync it is both better than native inference (+3.75\,pp) and cheaper (24 versus 35 forwards per decision).}}

\rev{GeoBoN, the closest candidate-allocation reference on both backbones, reports single-rollout RoboCasa baselines of 80.8\% for X-WAM and 66.3\% for Cosmos Policy and roughly two-point gains at $N=8$ \citep{zhao2026geobon}, which anchors the scale of our paired baselines.}

\rev{A small physical feasibility check on Mobile ALOHA \citep{fu2024mobilealoha}, applying the same inference change to one Mobile-ALOHA-tuned X-WAM checkpoint, is reported in Appendix~\ref{app:realrobot} (19/30 versus 22/30 successes).}

\section{Conclusion}

\looseness=-1
\rev{Asynchronous world--action models do not have a single test-time compute budget; they have two clocks. The commitment--evidence gap shows when the action is running ahead of its world evidence, the controlled schedule family shows how the preferred use of extra compute changes with that gap, and ReSync turns that structure into an inference rule with no new parameters, whose cost drops when the same world state is reached in fewer steps.} ReSync controls when the predicted future reaches the action; which resulting action changes to keep is a complementary question \citep{lin2026utilitywam}. \rev{\textbf{World prediction is valuable when it becomes available while the action can still change.}}

\clearpage
\section*{Reproducibility Statement}

\rev{Evaluation uses frozen RoboCasa (960), RoboTwin (300), and Cosmos (320) rollout panels, 384 paired decision states for the prediction-quality measurement, and 30 matched physical-robot trials per condition. The 8/40 and 12/60 operating points were recorded from the frozen rule before their local sweeps were evaluated. Appendix~\ref{app:stats} specifies statistical inference; Appendix~\ref{app:impl} gives schedules, pseudocode, controls, and the latency protocol. The supplementary video shows the mechanism and paired robot and simulator executions.}

\section*{Ethics Statement}

The work studies robot-manipulation inference and does not use human-subject
data. Physical-robot policy rollouts are executed in a bounded laboratory workspace
under operator supervision; operators provide resets and safety intervention
but do not teleoperate the evaluated trajectories.

\section*{AI Use Statement}

AI tools were used only for language polishing.

\nocite{black2024pi0,brohan2022rt1,brohan2023rt2,chi2023diffusionpolicy,jiang2022vima,khazatsky2024droid,kim2024openvla,mandlekar2023mimicgen,octo2024,oxe2023}
\nocite{lin2026wasd,lin2026batnav,lyu2026vlnevolution,wu2026dualanchoring}
\bibliographystyle{iclr2027_conference}
\setlength{\bibsep}{3pt plus 0.5pt minus 0.5pt}
\bibliography{refs_clean}
\newpage
\appendix
\pretocmd{\section}{\FloatBarrier}{}{}
\begin{center}{\Large\bfseries Appendix}\end{center}
\vspace{0.5em}

\section{Allocation Controls and Positioning}\label{app:alloc}

\begin{table}[!th]
  \centering
  \caption{\textbf{The neighboring methods act on different decision axes.} The final column describes the evidence state assumed or modeled at commitment.}
  \label{tab:position}
  \small
  \begin{tabular*}{\textwidth}{@{\extracolsep{\fill}}lllll@{}}
    \toprule
    \thead{Method} & \thead{Allocated object} & \thead{Signal} & \thead{Primary axis} & \thead{Evidence at commit} \\
    \midrule
    RoboMonkey & candidates       & VLM verifier      & candidate     & resolved (implicit) \\
    ADV        & drafts           & self-verification & candidate     & resolved (implicit) \\
    GeoBoN     & WAM trajectories & geometry          & candidate     & resolved (implicit) \\
    CAPS       & search           & SNR               & search depth  & unmodeled \\
    SANTS      & video steps      & scheduler         & stopping time & explicit noise \\
    NoiseGate  & latent timesteps & learned gate      & schedule      & explicit noise \\
    EVO        & cache refresh    & schedule search   & block$\times$time & unmodeled \\
    T-Stitch   & model capacity   & fixed switch      & timestep      & unmodeled \\
    ELASTIC    & seq./parallel    & meta-policy       & allocation    & unmodeled \\
    \textbf{ReSync} & \textbf{decision timing} & \textbf{schedule gap} & \textbf{commit time} & \textbf{explicit} \\
    \bottomrule
  \end{tabular*}
\end{table}

SANTS, NoiseGate, EVO, and ELASTIC correspond to adaptive stopping,
per-latent schedule gating, cache scheduling, and learned sequential/parallel
allocation, respectively
\citep{sun2026sants,huang2026noisegate,wang2026evo,li2026elastic}.

\begin{table}[!th]
\centering
\caption{\textbf{Allocation operating points on the updated paired panel.}}
\label{tab:grid}
\small
\begin{tabular*}{\textwidth}{@{\extracolsep{\fill}}lrrrrr@{}}
\toprule
Method & SR & $\Delta$pp & 95\% CI & Rescue/Break & Compute \\
\midrule
BASE & 80.31 & 0.00 & -- & -- & 1.00$\times$ \\
allocation $r25,N4$ & 82.71 & +2.40 & [+0.18,+4.61] & 64/41 & 1.75$\times$ \\
ReSync + allocation & 85.94 & +5.63 & [+3.17,+8.12] & 95/41 & 4.15$\times$ \\
\bottomrule
\end{tabular*}
\end{table}

\section{Simulation and Real-Robot Executions}\label{app:sim}

\subsection{Paired Simulation Trajectories}

\begin{figure}[!th]
  \centering
  \includegraphics[width=\textwidth]{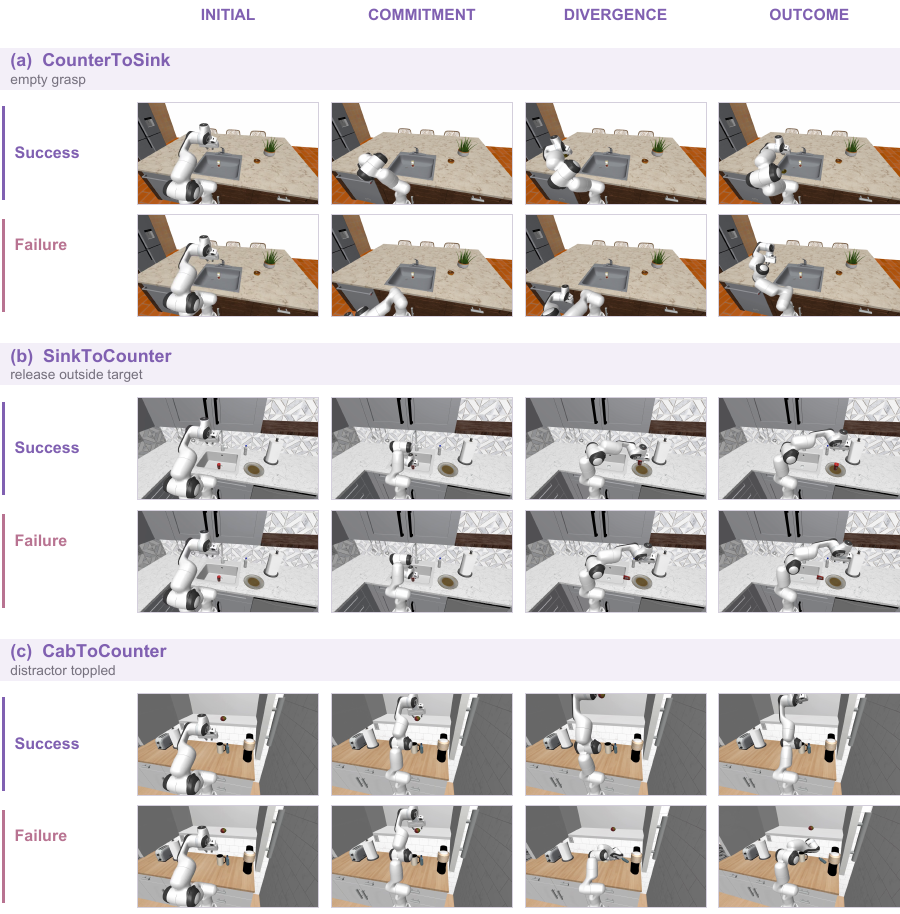}
  \caption{\textbf{Paired RoboCasa simulator executions.} Each pair starts from the same reset state and shows successful and failed executions. Frames are aligned at the initial state, gripper closure, divergence, and stable outcome.}
  \label{fig:sim}
  \label{fig:sim_robocasa}
\end{figure}

\begin{figure}[!th]
  \centering
  \includegraphics[width=\textwidth]{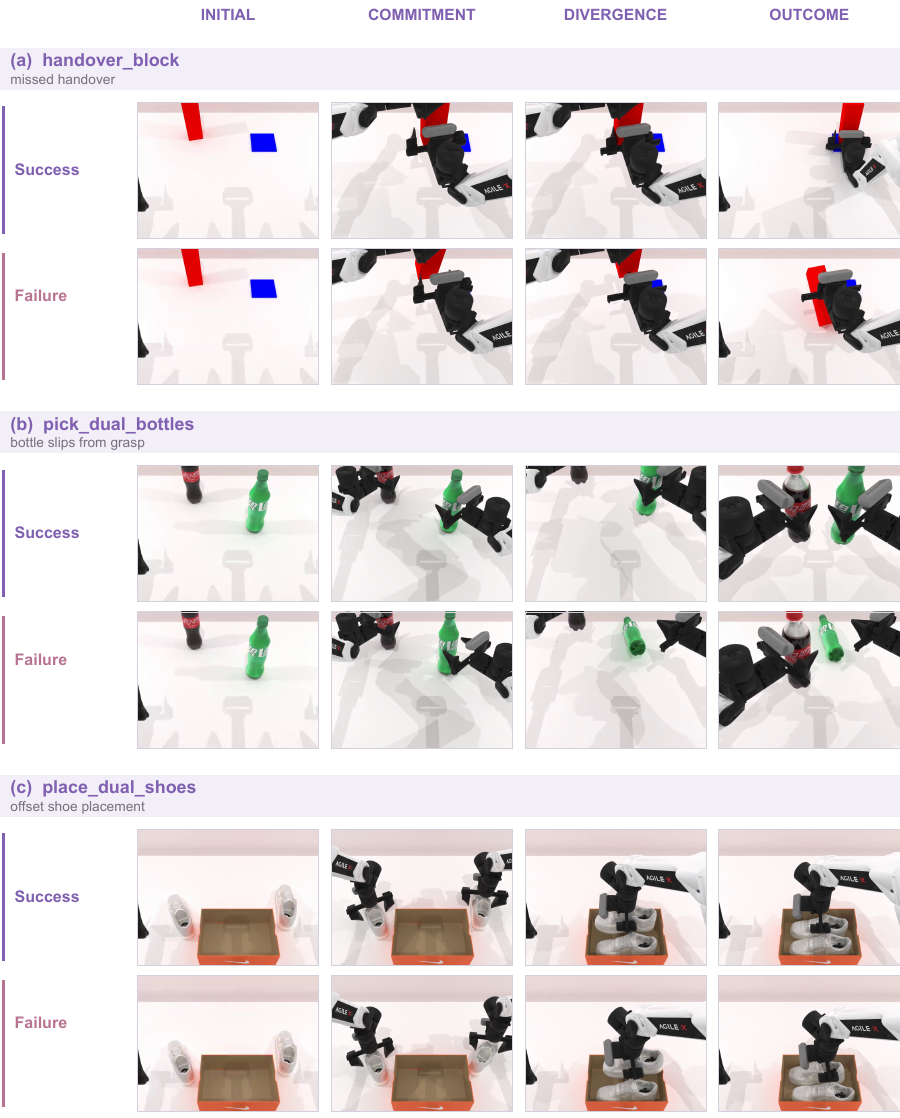}
  \caption{\textbf{Paired RoboTwin 2.0 simulator executions.} Each pair shares the initial state and shows successful and failed executions. Frames are aligned at gripper closure and subsequent task events. Source frames retain their native $640\times480$ resolution.}
  \label{fig:sim_robotwin}
\end{figure}

\begin{figure}[!th]
  \centering
  \includegraphics[width=\textwidth]{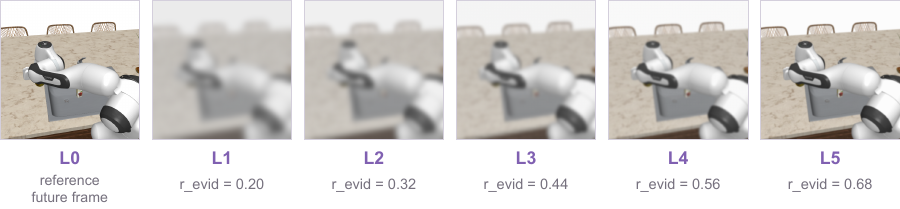}
  \caption{\textbf{Schematic schedule-progress ladder.} A rendered future frame
is degraded by controlled downsampling and blur to visualize the intuition behind
increasing schedule progress. L0 is the reference frame; L1--L5 are annotated
with schedule-progress values 0.20, 0.32, 0.44, 0.56, and 0.68. These panels are
an illustration of the proxy and are not model-decoded world states.}
  \label{fig:sim_focus}
\end{figure}

\subsection{Real-Robot Executions}
\label{app:realrobot}

The physical feasibility check uses one Mobile-ALOHA-tuned X-WAM checkpoint
for both BASE and ReSync. The same three-RGB/proprioceptive observation wrapper,
action-chunk controller, and checkpoint are used in both conditions; ReSync
changes only inference timing and receives no additional fine-tuning. We run 10
matched resets on each of three tasks. Cup placement changes from 6/10 to 7/10,
block placement from 6/10 to 8/10, and cluttered reaching from 7/10 to 7/10,
for an aggregate change from 19/30 (63.3\%) to 22/30 (73.3\%). The visualization
uses fixed episode indices rather than outcome-selected examples.

\begin{table}[!th]
\centering
\caption{\textbf{Physical-robot policy rollouts.} BASE and ReSync share the
same checkpoint, resets, and action interface.}
\label{tab:realrobot}
\small
\begin{tabular*}{0.72\textwidth}{@{\extracolsep{\fill}}lrrr@{}}
\toprule
Task & BASE & ReSync & $\Delta$ \\
\midrule
Cup placement       & 6/10 & 7/10 & +10\,pp \\
Block placement     & 6/10 & 8/10 & +20\,pp \\
Cluttered reaching  & 7/10 & 7/10 & 0\,pp \\
\midrule
Overall             & 19/30 & 22/30 & +10.0\,pp \\
\bottomrule
\end{tabular*}
\end{table}

\begin{figure}[!th]
  \centering
  \includegraphics[width=\textwidth]{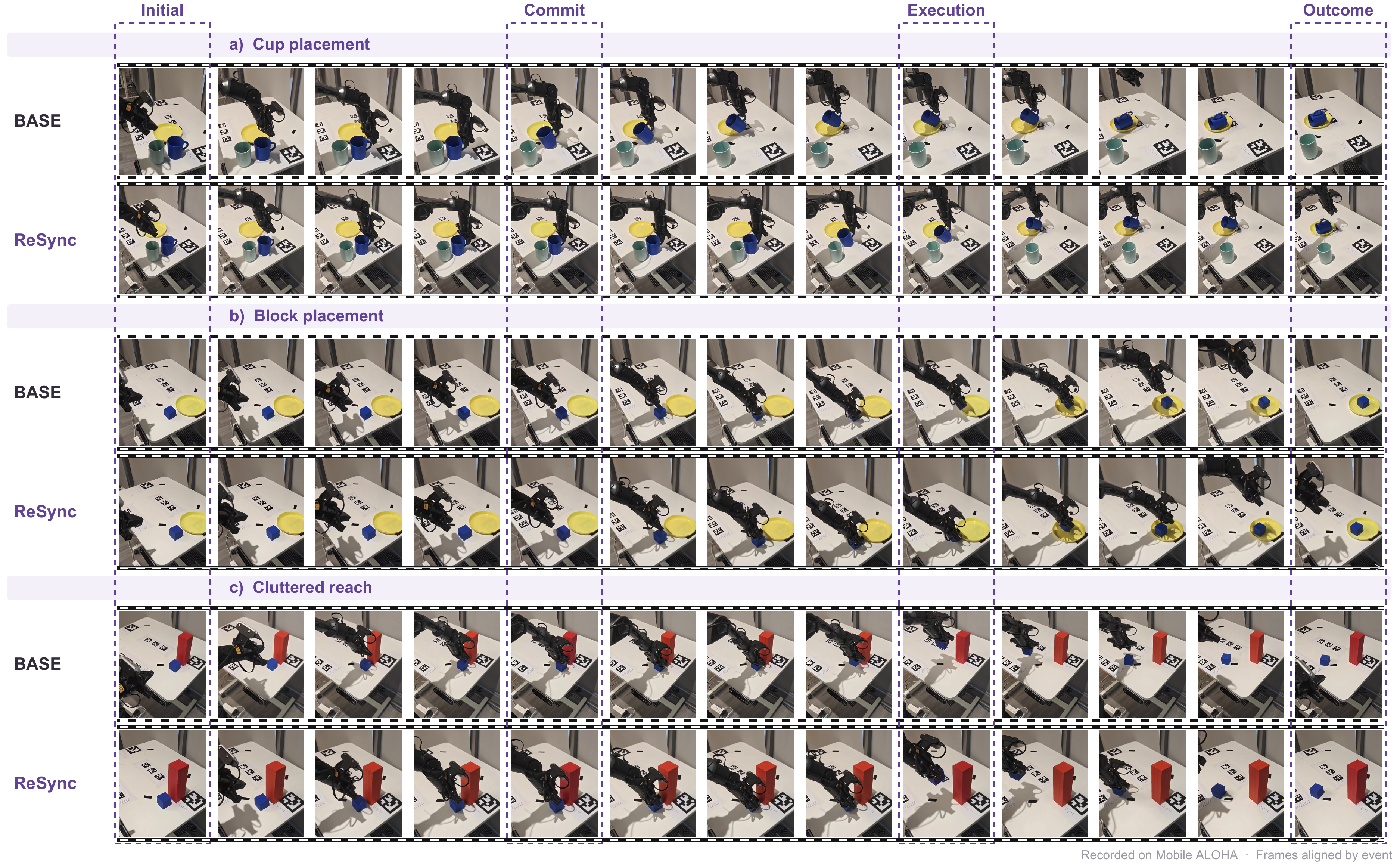}
  \caption{\textbf{Paired physical-robot policy rollouts.}
  BASE/ReSync pairs share task and reset conditions and are aligned at
  commitment and subsequent task events.}
  \label{fig:demo_real_appendix}
\end{figure}

Figure~\ref{fig:demo_real_appendix} uses the same fixed episode index for each
task rather than choosing examples by outcome. All 30 rollouts per condition
are retained in the experiment log.

\subsection{Learned Selector Stress Test}\label{app:selector}

The learned selector screen is separate from the fixed geometric scorer used in
Table~\ref{tab:gaplaw}. Here, a learned state--action-residual (SAR) selector
averages $-2.16$\,pp across 14 configurations (2/14 positive), while random
selection averages $-0.24$\,pp across 12 (5/12 positive). Across matched
schedule/depth pairs, random wins 9/12 (one-sided binomial $p=0.073$). The two selector results answer different questions:
the geometric scorer measures whether resolved candidate worlds become rankable under a fixed
label-free rule, while this learned screen tests whether a separate trained
state--action selector can exploit the same candidate bank.

\begin{figure}[!th]
  \centering
  \includegraphics[width=\textwidth]{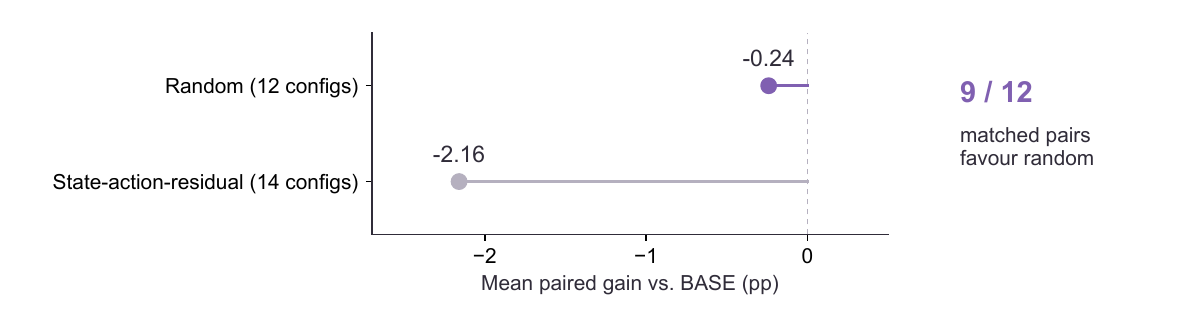}
  \caption{\textbf{The selector screen is directional rather than a headline result.} Mean paired deltas are shown for 14 SAR and 12 random-selector configurations. Random selection wins 9 of the 12 matched schedule/depth pairs.}
  \label{fig:selector}
\end{figure}

\section{In-Process Identifiability and Candidate Replacement}\label{app:inproc}

\subsection{Schedule and identifiability details}\label{app:prediction_details}

\lead{Controlled schedule family.} To isolate timing from architecture, we construct a controlled schedule family
from the same frozen X-WAM checkpoint. The action schedule remains fixed at ten
steps. We change only the world schedule length $T_O$. We refer to these
settings by their schedule lengths---\textsc{Sync-10}, \textsc{Mid-14},
\textsc{Mid-18}, \textsc{Mid-24}, \textsc{Mid-30}, and
\textsc{X-WAM (native)}---rather than as different models. The checkpoint,
observations, action head, candidate protocol, and evaluation panel are
identical across all six rows.

\lead{$\kappa$ protocol.} The identifiability screen uses 80 candidate sets
covering 22 tasks, with 32 candidates per set (2,560 candidate trajectories).
The representation family $\mathcal{R}$ contains four frozen scores
(consensus, state--action residual, cross-chunk consistency, and a random
control), four learned scores (absolute/relative targets with BCE or pairwise
training), visual world-latent scores based on $z_O$ and pooled relative forms,
and geometric centroid, nearest-neighbor, and median-distance scores. For each
of 10,000 within-set outcome permutations we recompute every score and the
familywise maximum AUC; $\mu_0$ is the mean of those maxima. Candidate-set
bootstrap intervals use the same set as the resampling unit. Native RoboCasa
gives $\kappa=0.0187$ (familywise $p=0.3408$, 95\% CI
$[-0.052,+0.086]$); the matched RoboTwin screen gives $\kappa=-0.079$
($p=0.8905$, 95\% CI $[-0.188,+0.031]$).

The $N=1$ oracle value of 53.8\% belongs to this separate 80-set candidate-bank
screen. The main 80.31\% BASE result is measured on the frozen 960-rollout,
24-task RoboCasa policy panel, so the two numbers estimate different
quantities under different evaluation protocols.

\noindent\textbf{A falsifiable consequence of unresolved evidence.}
Consider a masked-evidence channel
\[
    P_y^r = rP_y^\star + (1-r)Q,
\]
where $P_y^\star$ denotes fully resolved evidence for utility class $y$ and the
unresolved component $Q$ is independent of that class. Then
\[
    \mathrm{TV}(P_1^r,P_0^r)
      = r\,\mathrm{TV}(P_1^\star,P_0^\star)
      \le r.
\]
Unresolved mass contracts the discrimination available at commitment in
direct proportion to the resolved fraction. This identity motivates a directional diagnostic for the real WAM: with
candidates and scorer fixed, advancing only the world should increase within-set
identifiability, while widening the gap should reduce the value of candidate
discrimination.

\lead{Candidate-bank and evidence controls.} The candidate bank has ample oracle headroom: nested-prefix oracle success rises
from 53.8\% at $N=1$ to 77.6\% at $N=32$. Yet native X-WAM commitment exposes
almost none of it to a deployment-time selector. Across frozen, learned, visual,
and geometric representations, within-set discrimination stays near chance
($\kappa=0.0187$ on RoboCasa; $-0.079$ on the matched RoboTwin subset), and
replacing one quarter of the candidate pool substantially reorders rankings
while barely moving closed-loop success.

The matched world-only intervention in Section~\ref{sec:pred1} changes only
when the decision reads the world. Varying $T_O$ from the opposite direction
produces the same ordering: larger commitment--evidence gaps reduce selector
value and increase the return to evidence advancement
(Figure~\ref{fig:gaplaw}, Table~\ref{tab:gaplaw}).

\lead{Schedule length and native quality.} The BASE column reveals the second half of the story. Native success is itself
non-monotonic in $T_O$: aggressively shortening the world schedule closes the
gap but sacrifices generation quality, while a long world trajectory preserves
prediction capacity but leaves more of it unresolved at commitment. ReSync
keeps the rich trajectory and reallocates only its useful inference-time
portion before commitment.

\lead{Routing-map consistency check.} This check audits which routed bins the
native sampler actually visits; it is an implementation consistency check, not
an independent predictive result. A small frozen-backbone coupling adapter
routes with
$k(t)=\lfloor(1-t/1000)\,10\rfloor$. The ten executed X-WAM action timesteps
map to $[0,0,0,0,1,1,2,3,4,6]$, giving reachable support
$\{0,1,2,3,4,6\}$ and dead support $\{5,7,8,9\}$ \emph{before training}.

\rev{The executed routing map matches 10/10 bins across all three seeds (Figure~\ref{fig:arch}c).} The
non-monotonic hole is a useful implementation fingerprint: $k=5$ stays at
initialization while $k=6$ learns. The dead-$k5$/live-$k6$ pattern follows
directly from the routed timestep map.

\lead{Preservation and intervention.} Reachability alone is not sufficient:
modifying the reachable bins degrades the competent native path. These controls
make train-time coupling modification unattractive and motivate a
parameter-free temporal intervention. Full coupling-adapter optimization,
native-path success change, variance trajectories, and seedwise controls are in
Appendix~\ref{app:impl}.

\bluechange{\lead{Matched-budget selector.} The same-budget and $3.40\times$ selector columns of Table~\ref{tab:gaplaw} use the fixed label-free scorer with $N=4$ candidates applied to a fixed random fraction of decisions, chosen so that average compute equals the target budget: 6.67\%, 13.33\%, 26.67\%, 40\%, and 80\% of decisions for 1.20$\times$ to 3.40$\times$. Absolute success rates: same budget 76.15, 79.69, 81.46, 81.77, 81.04, 80.52; common $3.40\times$ 78.33, 81.56, 82.81, 82.19, 81.25, 80.52; ReSync 76.36, 80.31, 82.71, 83.96, 84.27, 84.79 (Sync-10 to native).}

\subsection{Additional identifiability controls}

A label-free world-particle geometry screen evaluates six geometric scores at all ten executed action steps. Every within-set AUC remains near chance and the familywise max-statistic gives $p=1.0$.

\begin{table}[!th]
  \centering
  \caption{\textbf{In-process geometry remains weak at every executed action position.} Values are within-set AUC.}
  \label{tab:anssmc}
  \small
  \begin{tabular*}{\textwidth}{@{\extracolsep{\fill}}lrrrrrrrrrr@{}}
    \toprule
    \thead{Metric} & \thead{$t_0$} & \thead{$t_1$} & \thead{$t_2$} & \thead{$t_3$} & \thead{$t_4$} & \thead{$t_5$} & \thead{$t_6$} & \thead{$t_7$} & \thead{$t_8$} & \thead{$t_9$} \\
    \midrule
    kNN density       & .435 & .424 & .428 & .433 & .436 & .426 & .431 & .431 & .428 & .432 \\
    Mahalanobis       & .433 & .483 & .505 & .479 & .462 & .460 & .453 & .493 & .500 & .471 \\
    KDE               & .516 & .516 & .513 & .514 & .516 & .516 & .516 & .516 & .516 & .516 \\
    Dist.-center      & .489 & .489 & .486 & .486 & .486 & .486 & .489 & .486 & .486 & .486 \\
    Local agree.      & .482 & .491 & .485 & .497 & .488 & .491 & .488 & .482 & .483 & .494 \\
    Full dist.-center & .449 & .449 & .449 & .449 & .449 & .449 & .449 & .449 & .449 & .449 \\
    \bottomrule
  \end{tabular*}
\end{table}

\lead{Candidate replacement.} Replacing one quarter of the candidate pool reorders 38.7\% of cases; 50\% replacement reorders 61.4\%. The 25\% intervention moves success by +0.52\,pp, 95\% CI $[-1.22,+2.31]$, with OR$\,\approx1.08$ and $\phi\approx0.013$.

\begin{proposition}[Masked-evidence contraction]
If unresolved evidence is class-independent,
$P^r_y=rP^\star_y+(1-r)Q$, then
\[
\mathrm{TV}(P^r_1,P^r_0)
=r\,\mathrm{TV}(P^\star_1,P^\star_0)\le r.
\]
\end{proposition}
\noindent The proposition is a compact masked-evidence identity used to motivate
the identifiability test in Section~\ref{sec:pred1}; the empirical question is
whether the real WAM follows the same direction when only the world clock is
advanced.

\section{ReSync Implementation, Schedule Rule, and Controls}\label{app:impl}

\lead{Implementation details.} ReSync is always-on once the schedule-derived
operating point is selected. ReSync+allocation gates only residual candidate
continuations after the shared ReSync prefix; it does not alter the ReSync
placement or world-advance rule.

\lead{Index convention.} The schedule-rule index $j$ denotes the last completed action update; the placement-scan index $t_i=j+1$ denotes the next action update. With $J=10$, the remaining budget is $R=J-1-j=10-t_i$, including the update at $t_i$. Therefore $j^*=6$ corresponds to $t_i=7$ and leaves three updates, whereas $j=8$ leaves one.

\lead{Cosmos convention.} Cosmos uses a different training construction from
X-WAM, so its transfer test is defined on the constructed inference interface
itself. We use a 14-update action stream against the 35-step future-state
schedule and carry over the same recovery convention:
$J_C-1-j_C\ge R_{\min}=3$, giving $j_C=10$. The interface fixes
$M_C=10$ world-only forwards before evaluation. ReSync caches after action
update 10, performs those ten world-only forwards while restoring the cached
action state, and completes native action updates 11--13. BASE and ReSync both
execute the same 14 action updates, and compute is $1+10/14=1.71\times$.

\lead{Native schedule and $\psi$.} The runner uses independent FlowUniPC schedules with 50 world and ten action steps and time-shift 5. During the first ten joint iterations it indexes world and action schedules with the same loop index, so $\psi(j)=j$. The action timesteps are
\[
  [999, 978, 952, 920, 882, 833, 768, 681, 555, 356].
\]
The 50-step world schedule is
\begin{align*}
  [&999, 995, 991, 987, 982, 978, 973, 968, 963, 957, 952, 946, 940, 934, 927, 920, 913, 906, 898, 890,\\
   &882, 873, 863, 854, 843, 833, 821, 809, 796, 783, 768, 753, 737, 720, 701, 681, 660, 636, 611, 584,\\
   &555, 522, 487, 448, 405, 356, 302, 241, 172, 92].
\end{align*}
Following X-WAM's joint-timestep sampling (Sec.~3.3 of \citet{guo2026xwam}), we define the represented joint noisy
regime by $t_O\ge t_a$; hence $M_{\max}(j)=\max\{m: t_O^{\psi(j)+m}\ge t_a^j\}$.

\begin{algorithm}[!th]
\color{black}
\captionsetup{labelfont={color=black}}
\caption{\rev{ReSync at inference (one policy decision).}}
\label{alg:resync}
\begin{algorithmic}[1]
\Require frozen WAM; native action schedule $(t_a^j)_{j<J}$; native world schedule $(t_O^i)_{i<T_O}$; $R_{\min}$
\State $j^\star \gets \max\{\,j\in\mathcal{S}_{\mathrm{exec}} : J-1-j\ge R_{\min}\}$;\quad $M\gets M_{\max}(j^\star)$
\State Run native joint denoising through action update $j^\star$
\State Cache the action latent and its timestep
\For{$m=1,\dots,M$} \Comment{lighter variants: every $(M/K)$-th world timestep}
  \State One shared-DiT forward; keep the world update; restore the cached action
\EndFor
\State Resume native joint denoising for the remaining $J-1-j^\star$ action updates
\State \Return the committed action chunk
\end{algorithmic}
\end{algorithm}

\begin{table}[!th]
  \centering
  \caption{\textbf{The schedule makes the X-WAM operating point directly auditable.} $R_{\min}=3$ removes $j=7,8$ and leaves $j^\star=6$ as the latest safe maximizer.}
  \label{tab:schedrule}
  \small
  \begin{tabular*}{\textwidth}{@{\extracolsep{\fill}}rrrrrr@{}}
    \toprule
    \thead{$j$} & \thead{$t_a^j$} & \thead{$\psi(j)$} & \thead{$t_O^{\psi(j)}$} & \thead{$M_{\max}$} & \thead{recovery} \\
    \midrule
    0 & 999 & 0 & 999 & 0 & 9 \\
    1 & 978 & 1 & 995 & 4 & 8 \\
    2 & 952 & 2 & 991 & 8 & 7 \\
    3 & 920 & 3 & 987 & 12 & 6 \\
    4 & 882 & 4 & 982 & 16 & 5 \\
    5 & 833 & 5 & 978 & 20 & 4 \\
    \textbf{6} & \textbf{768} & \textbf{6} & \textbf{973} & \textbf{24} & \textbf{3} \\
    7 & 681 & 7 & 968 & 28 & 2 \\
    8 & 555 & 8 & 963 & 32 & 1 \\
    \bottomrule
  \end{tabular*}
\end{table}

\lead{Frozen-rule schedule pairs.} Applying the same scheduler construction and
the already fixed $R_{\min}=3$ to two unused schedule pairs gives
$(J,T_O)=(8,40)\Rightarrow j^\star=4,\;M^\star=16$ and
$(12,60)\Rightarrow j^\star=8,\;M^\star=32$. These operating points were
recorded before the corresponding neighbor and warm-start evaluations in
Table~\ref{tab:heldout_schedules}; no value from those sweeps enters the rule.

The compute--quality tradeoff is explicit. On one H800, native X-WAM takes
2.68\,s per policy decision (0.37\,Hz), $M=8$ takes 4.74\,s (0.21\,Hz), and
$M=24$ takes 8.97\,s (0.11\,Hz). These are policy-decision intervals, not low-level servo periods. \rev{The public X-WAM RoboCasa evaluator blocks on the policy response and then executes the returned 32-action chunk before issuing the next query \citep{guo2026xwam}.}
Applications that value faster replanning can move left on the same curve rather
than changing the rule itself.

\lead{Latency and overshoot.} Wall-clock growth is slightly sub-linear relative to forward-equivalent compute because batching, model dispatch, and compiled-graph overhead are partly fixed. The $M=32,40$ rows intentionally exceed the $M_{\max}(6)=24$ training-support boundary and expose the descending side of the interior optimum.

\begin{table}[!th]
\centering
\caption{\textbf{World advancement peaks at the training-supported boundary.}
Wall-clock measurements use one NVIDIA H800 80GB GPU with identical batching
and compilation state; $M>24$ deliberately crosses joint training support.}
\label{tab:curve}
\small
\begin{tabular*}{\textwidth}{@{\extracolsep{\fill}}rrrrrrr@{}}
\toprule
$M$ & Schedule progress proxy & SR & $\Delta$pp & Compute & Latency / decision & Effective Hz \\
\midrule
0  & 0.20 & 80.31 & 0.00  & 1.00$\times$ & 2.68\,s & 0.37 \\
8  & 0.36 & 81.67 & +1.35 & 1.80$\times$ & 4.74\,s & 0.21 \\
16 & 0.52 & 83.33 & +3.02 & 2.60$\times$ & 6.83\,s & 0.15 \\
24 & 0.68 & 84.79 & +4.48 & 3.40$\times$ & 8.97\,s & 0.11 \\
32 & 0.84 & 83.96 & +3.65 & 4.20$\times$ & 11.14\,s & 0.09 \\
40 & 1.00 & 82.29 & +1.98 & 5.00$\times$ & 13.32\,s & 0.08 \\
\bottomrule
\end{tabular*}
\end{table}

\bluechange{\lead{Lighter variants.} The $K=2$, $K=4$, $K=8$ and $K=12$ variants keep $j^\star=6$ and the same world target as $M=24$, and take every 12th, 6th, 3rd or 2nd of the 24 inserted world timesteps. 95\% cluster-bootstrap CIs relative to BASE: $K=2$ [$-0.72$,+3.16], $K=4$ [+0.43,+4.80], $K=8$ [+1.34,+5.74], $K=12$ [+1.69,+6.13], $K=8$ + allocation [+2.42,+7.16].}

\begin{table}[!th]
\centering\color{black}
\captionsetup{labelfont={rm,color=blue},textfont={color=blue}}
\caption{\bluechange{\textbf{Lighter-variant paired results on RoboCasa.} 960 paired rollouts in 192 physical-state clusters; 95\% CIs use cluster-level paired bootstrap.}}
\label{tab:lighter_ci}
\small
\begin{tabular*}{\textwidth}{@{\extracolsep{\fill}}lrrr@{}}
\toprule
Method & SR (\%) & $\Delta$ vs BASE (pp) & 95\% CI (pp) \\
\midrule
ReSync Coarse $K=2$ & 81.56 & +1.25 & [$-0.72$,+3.16] \\
ReSync Coarse $K=4$ & 82.92 & +2.60 & [+0.43,+4.80] \\
ReSync Coarse $K=8$ & 83.85 & +3.54 & [+1.34,+5.74] \\
ReSync Coarse $K=12$ & 84.27 & +3.96 & [+1.69,+6.13] \\
ReSync $K=8$ + Allocation & 85.10 & +4.79 & [+2.42,+7.16] \\
\bottomrule
\end{tabular*}
\end{table}

\bluechange{\lead{Prediction-quality measurement.} For each of 192 clusters we take two decision states (the first decision and the midpoint decision of a BASE rollout), run each variant with the same noise seed, decode the world prediction at commitment, execute the committed 32-action chunk in the simulator, and compare predicted and observed frames at matching times. LPIPS uses the full frame; task-region error uses simulator segmentation of the gripper and target object and is normalized by the native value.}

\bluechange{\lead{Optimized inference and timing replay.} The optimized implementation compiles the model with \texttt{torch.compile} (TorchInductor), replays it with CUDA graphs under static shapes, uses FlashAttention-3, and splits the DiT token sequence across four H800s with DeepSpeed Ulysses, overlapping the NCCL all-to-all communication with computation. Every method runs in BF16, the same numerical precision as the research code; display-only video decoding that does not feed action generation is skipped, and neither quantization (FP8, NVFP4) nor approximate KV caching is used. Latency is measured end to end, from observation input to action-chunk output, over 384 decisions per method after 32 warm-up decisions, on the same GPU type for both implementations; on a single H800 the same implementation takes 1.85\,s for native X-WAM and 3.27\,s for $K=8$ (P50). Running both implementations on 384 decision states with the same observation and noise seed gives nearly identical 32-action chunks: for every method in Table~\ref{tab:async}, the median action RMSE is at most 0.0014, the 95th-percentile RMSE at most 0.0033, and the maximum element error at most 0.0082 in normalized action space. The timing replay therefore runs the research implementation and injects the measured latency distribution of either implementation in simulated time. While a chunk executes, the next one is computed; if it has not arrived when the chunk ends, the robot holds its last command, and the new chunk is anchored to its actual arrival time. Episodes contain 7 to 15 decisions (median 12). Under optimized timing, every method except $M=24$ loses the same two episodes relative to blocking execution.}

\begin{table}[!th]
\centering\color{black}
\caption{\bluechange{\textbf{RoboCasa asynchronous execution with measured latency.} The full 960-rollout panel is replayed at 20\,Hz with a 32-action chunk and a 1.6\,s execution window, injecting the measured per-decision latency distribution in simulated time. Blocking: the simulator waits for every decision, as in the main text. Pause: mean hold per decision. Episode: median wall-clock time.}}
\label{tab:async}
\small
\begin{tabular*}{\textwidth}{@{\extracolsep{\fill}}lrrrrrrr@{}}
\toprule
& & \multicolumn{3}{c}{Research code, 1 GPU} & \multicolumn{3}{c}{Optimized, 4 GPUs} \\
\cmidrule(lr){3-5}\cmidrule(lr){6-8}
Method & Blocking SR & Async SR & Pause & Episode & Async SR & Pause & Episode \\
\midrule
X-WAM (native)      & 80.31 & 79.69 & 1.10\,s & 33.9\,s  & 80.10 & 0.000\,s     & 19.9\,s \\
ReSync, $K=2$       & 81.56 & 80.83 & 1.62\,s & 39.6\,s  & 81.35 & 0.000\,s     & 20.1\,s \\
ReSync, $K=4$       & 82.92 & 82.19 & 2.12\,s & 45.6\,s  & 82.71 & $<$0.001\,s  & 20.2\,s \\
ReSync, $K=8$       & 83.85 & 83.23 & 3.16\,s & 58.6\,s  & 83.65 & 0.005\,s     & 20.5\,s \\
ReSync, $M=24$      & 84.79 & 84.06 & 7.41\,s & 109.4\,s & 84.17 & 1.03\,s      & 32.5\,s \\
allocation $r25,N4$ & 82.71 & 81.88 & 3.25\,s & 58.1\,s  & 82.50 & 0.43\,s      & 24.7\,s \\
\bottomrule
\end{tabular*}
\end{table}

\lead{Preservation control.} The coupling-adapter support test separates preservation from reachability. Uniform capacity yields native-path success change $(-0.83,-1.07,-2.29)$\,pp across seeds; reachable-only and dead-only closed-loop reallocations yield $(-0.46,-0.71,-1.18)$\,pp and $(+0.04,-0.03,+0.02)$\,pp respectively. The distinct metrics are kept separate because they answer different questions: whether the native path is preserved, and whether train-time reallocation remains a viable policy intervention.

\begin{figure}[!th]
  \centering
  \includegraphics[width=0.5\textwidth]{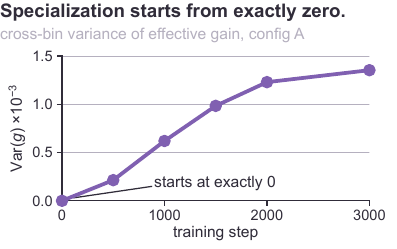}
  \caption{\textbf{Specialization is learned, not inherited.} Cross-bin variance of the effective gain during the coupling-adapter support test. It is zero at initialization and rises monotonically; the unvisited bins remain at 0.50.}
  \label{fig:var}
\end{figure}

\section{Reproducibility and Statistics}\label{app:stats}

\begin{table}[!th]
\centering\color{black}
\caption{\rev{\textbf{Paired rescue and break counts on the 960-rollout panel.}}}
\label{tab:rescue}
\small
\renewcommand{\arraystretch}{0.95}
\begin{tabular*}{\textwidth}{@{\extracolsep{\fill}}lrrrr@{}}
\toprule
Method & Compute & SR & $\Delta$pp & Rescue/Break \\
\midrule
allocation $r25,N4$ & 1.75$\times$ & 82.71 & +2.40 & 64/41 \\
wait-only & 3.40$\times$ & 80.52 & +0.21 & 43/41 \\
matched resampling & 3.40$\times$ & 82.08 & +1.77 & 66/49 \\
Action-34 & 3.40$\times$ & 81.56 & +1.25 & 50/38 \\
Sync-34 & 3.40$\times$ & 82.29 & +1.98 & 57/38 \\
world warm-start & 3.40$\times$ & 82.92 & +2.60 & 61/36 \\
ReSync & 3.40$\times$ & 84.79 & +4.48 & 82/39 \\
Sync-50 & 5.00$\times$ & 83.23 & +2.92 & 68/40 \\
ReSync+allocation & 4.15$\times$ & 85.94 & +5.63 & 95/41 \\
\bottomrule
\end{tabular*}
\end{table}

The evaluation grid, selector, trigger masks, seed contract, paired estimands,
cluster key, and tests are fixed before evaluating the 960-case panel. The
panel is 24 tasks $\times$ eight state IDs $\times$ five rollout IDs. Cluster
bootstrap operates on 192 physical states; exact McNemar uses paired
rescue/break counts. The RoboTwin ReSync panel is 60 clusters $\times$ five
replicates.

\lead{Schedule-shaping contrasts.} A random-placement schedule is the direct
reference for early-, late-, and decayed-placement variants. Cluster-bootstrap
CI is the primary inference unit; case-wise exact McNemar is secondary.
Table~\ref{tab:schedule_contrasts} reports the four contrasts.

\begin{table}[!th]
  \centering
  \caption{\textbf{Schedule shaping changes point estimates, not clustered inference.} McNemar is a secondary diagnostic.}
  \label{tab:schedule_contrasts}
  \small
  \begin{tabular*}{\textwidth}{@{\extracolsep{\fill}}lrr@{\hspace{12pt}}r@{}}
    \toprule
    \thead{Contrast} & \thead{$\Delta$pp} & \thead{cluster-bootstrap 95\% CI} & \thead{McNemar $p$} \\
    \midrule
    random placement vs BASE    & +0.94  & [-1.16,+3.07]   & 0.39 \\
    early vs random & +0.28 & [-1.51,+2.11] & 0.76 \\
    late vs random  & -0.86  & [-2.84,+1.13]   & 0.35 \\
    decay vs random & +1.31 & [-0.58,+3.22] & 0.17 \\
    \bottomrule
  \end{tabular*}
\end{table}

\lead{Historical protocols.} Published X-WAM reports 79.2\%; an early 24-task$\times$40-episode reproduction gives 79.79\%; the frozen clustered paired baseline used here is 80.31\%. These are separate protocols and are not merged.

\begin{table}[!th]
\centering
\color{black}
\captionsetup{labelfont={rm,color=blue},textfont={color=blue}}
\caption{\bluechange{\textbf{Equal-budget contrasts against ReSync with Holm correction.}}}
\label{tab:holm}
\small
\begin{tabular*}{\textwidth}{@{\extracolsep{\fill}}p{0.39\textwidth}rrrr@{}}
\toprule
Contrast & ReSync advantage (pp) & 95\% CI & bootstrap $p$ & Holm $p$ \\
\midrule
vs wait-only          & +4.27 & [+2.40,+6.25] & 0.0004 & 0.0020 \\
vs Action-34          & +3.23 & [+1.28,+5.19] & 0.0013 & 0.0052 \\
vs matched resampling & +2.71 & [+0.86,+4.57] & 0.0042 & 0.0126 \\
vs Sync-34            & +2.50 & [+0.61,+4.39] & 0.0096 & 0.0192 \\
vs world warm-start   & +1.87 & [+0.09,+3.68] & 0.0410 & 0.0410 \\
\midrule
warm-start, pooled over 10/50, 8/40, 12/60 & +1.63 & [+0.57,+2.73] & 0.004 & -- \\
vs Sync-50 ($5.00\times$) & +1.56 & [-0.28,+3.42] & -- & -- \\
\bottomrule
\end{tabular*}
\end{table}

\section{Evidence Lineage}\label{app:lineage}

The experiment records retain the frozen 960-case panel, case identifiers,
task/state/rollout mapping, candidate seeds, candidate-replacement controls,
coupling-adapter checks, placement scans, selector screens, ReSync ablations,
and the pre-evaluation operating-point records for the 8/40 and 12/60 schedule
pairs. Quantitative tables are generated from the corresponding frozen
evaluation records. The supplementary package contains the narrated video and a
short README.

\lead{Training-support extension.} ReSync's ceiling follows joint timestep support represented during training. Widening that support directly expands the admissible evidence-advancement region.

\end{document}